\documentclass[10pt,twocolumn,letterpaper]{article}

\usepackage{iccv}
\usepackage{times}
\usepackage{epsfig}
\usepackage{graphicx}
\usepackage{amsmath}
\usepackage{amssymb}

\makeatletter
\@namedef{ver@everyshi.sty}{}
\makeatother
\usepackage{booktabs}
\usepackage{multirow}
\usepackage{pgfplots}
\usepackage[export]{adjustbox}
\pgfplotsset{compat=1.17}
\usepackage{siunitx}
\DeclareSIUnit[number-unit-product = ]\k{k}
\usepackage[mode=buildnew]{standalone}
\usepackage{calc}
\usepackage{xr-hyper}
\usepackage{filecontents}
\usepackage{colortbl}
\usepackage{pgfplotstable}
\usetikzlibrary{pgfplots.colorbrewer}
\usepackage[noadjust,nocompress]{cite}

\newcommand{\ra}[1]{\renewcommand{\arraystretch}{#1}}
\newcommand{\PAR}[1]{\vskip2pt \noindent{\bf #1}}
\newcommand{\textover}[3][l]{%
 \makebox[\widthof{#3}][#1]{#2}%
}

\usepackage[page]{appendix}

\newcommand{\beginappendixa}{%
        \setcounter{table}{0}
        \renewcommand{\thetable}{A-\arabic{table}}%
        \setcounter{figure}{0}
        \renewcommand{\thefigure}{A-\arabic{figure}}%
     }
\newcommand{\beginappendixb}{%
        \setcounter{table}{0}
        \renewcommand{\thetable}{B-\arabic{table}}%
        \setcounter{figure}{0}
        \renewcommand{\thefigure}{B-\arabic{figure}}%
     }
\newcommand{\beginappendixc}{%
        \setcounter{table}{0}
        \renewcommand{\thetable}{C-\arabic{table}}%
        \setcounter{figure}{0}
        \renewcommand{\thefigure}{C-\arabic{figure}}%
     }
\newcommand{\beginappendixd}{%
        \setcounter{table}{0}
        \renewcommand{\thetable}{D-\arabic{table}}%
        \setcounter{figure}{0}
        \renewcommand{\thefigure}{D-\arabic{figure}}%
     }
\newcommand{\beginappendixe}{%
        \setcounter{table}{0}
        \renewcommand{\thetable}{E-\arabic{table}}%
        \setcounter{figure}{0}
        \renewcommand{\thefigure}{E-\arabic{figure}}%
     }
\newcommand{\beginappendixf}{%
        \setcounter{table}{0}
        \renewcommand{\thetable}{F-\arabic{table}}%
        \setcounter{figure}{0}
        \renewcommand{\thefigure}{F-\arabic{figure}}%
     }

\usepackage[pagebackref=true,breaklinks=true,letterpaper=true,colorlinks,bookmarks=false]{hyperref}

\iccvfinalcopy 

\def\iccvPaperID{3271} 

\ificcvfinal\fi

\begin{document}

\title{Contrastive Model Adaptation for Cross-Condition Robustness\\in Semantic Segmentation}

\author{David Bruggemann
\qquad Christos Sakaridis
\qquad Tim Brödermann
\qquad Luc Van Gool \\
ETH Zurich, Switzerland \\
{\tt\small \{brdavid, csakarid, timbr, vangool\}@vision.ee.ethz.ch}}

\maketitle
\ificcvfinal\thispagestyle{empty}\fi

\begin{abstract}
Standard unsupervised domain adaptation methods adapt models from a source to a target domain using labeled source data and unlabeled target data jointly.
In model adaptation, on the other hand, access to the labeled source data is prohibited, i.e., only the source-trained model and unlabeled target data are available.
We investigate normal-to-adverse condition model adaptation for semantic segmentation, whereby image-level correspondences are available in the target domain.
The target set consists of unlabeled pairs of adverse- and normal-condition street images taken at GPS-matched locations.
Our method\textemdash CMA\textemdash leverages such image pairs to learn condition-invariant features via contrastive learning.
In particular, CMA encourages features in the embedding space to be grouped according to their condition-invariant semantic content and not according to the condition under which respective inputs are captured.
To obtain accurate cross-domain semantic correspondences, we warp the normal image to the viewpoint of the adverse image and leverage warp-confidence scores to create robust, aggregated features.
With this approach, we achieve state-of-the-art semantic segmentation performance for model adaptation on several normal-to-adverse adaptation benchmarks, such as ACDC and Dark Zurich.
We also evaluate CMA on a newly procured adverse-condition generalization benchmark and report favorable results compared to standard unsupervised domain adaptation methods, despite the comparative handicap of CMA due to source data inaccessibility.
Code is available at \url{https://github.com/brdav/cma}.

\end{abstract}


\section{Introduction}

\begin{figure}
\centering
    \includegraphics[width=\linewidth]{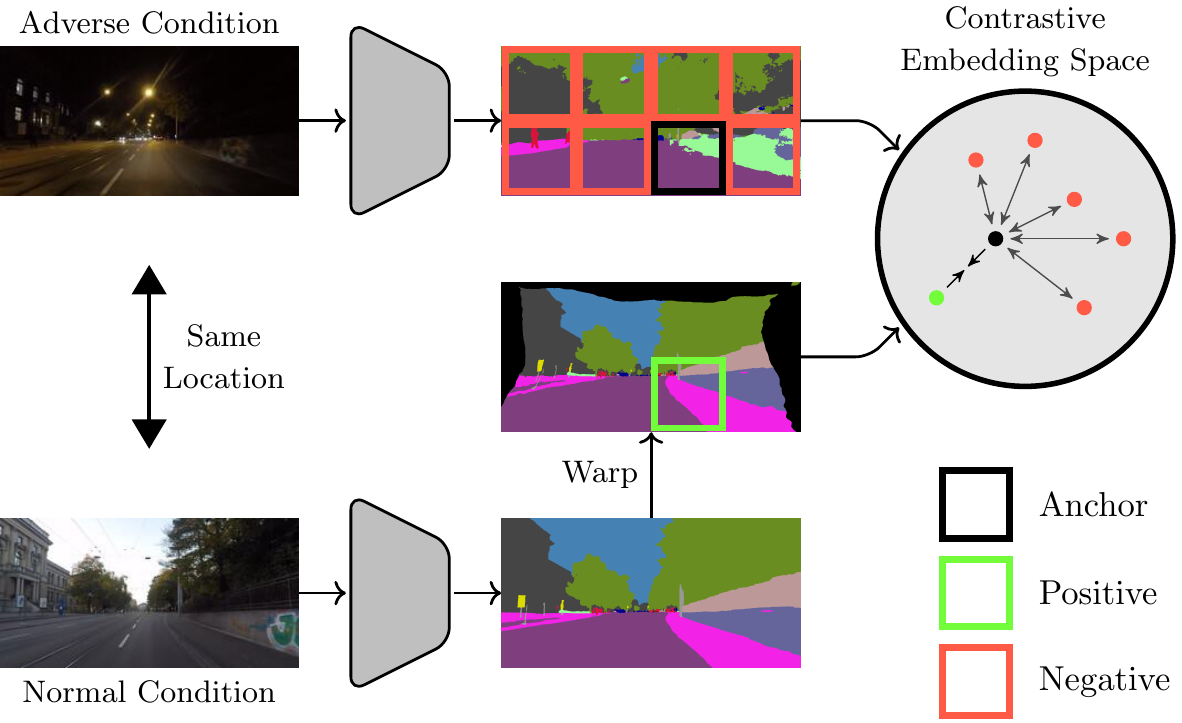}
    \caption{CMA exploits image-level correspondences to learn condition-invariant features. Two images of the same location (but captured under different visual conditions) are encoded, and the normal-condition image features are warped to get spatially aligned with the adverse-condition image features. Our contrastive loss then creates an embedding space where patches of adverse features (black) are \emph{closer to their corresponding normal patches} (green) than to other adverse patches (red).}
    \label{fig:teaser}
    \vspace{-0.2em}
\end{figure}

Adverse visual conditions, such as fog, heavy rain, or snowfall, represent a challenge for autonomous systems expected to navigate ``in the wild''.
To achieve full autonomy, systems require perception algorithms that perform robustly in every condition.
However, due to their infrequent occurrence, inclement weather conditions  are often underrepresented in common, finely annotated outdoor datasets (\eg, BDD100K~\cite{yu2020bdd100k} or Mapillary Vistas~\cite{neuhold2017mapillary}).
As a result, state-of-the-art recognition methods are biased towards ``normal'' visual conditions (\ie, daytime and clear weather), which causes them to fail for edge cases.
Furthermore\textemdash in particular for detailed, pixel-level tasks like semantic segmentation\textemdash high-quality annotations for adverse-condition images are difficult and expensive to obtain.
In fact, they require specialized annotation protocols due to ambiguities arising from aleatoric uncertainty~\cite{sakaridis2021acdc}.
To bypass these issues, researchers have investigated 
unsupervised domain adaptation (UDA) from normal to adverse conditions as an alternative to full supervision~\cite{hoffman2016fcns,wulfmeier2017addressing,chen2018domain,dai2018dark,sakaridis2023condition}, where a model is jointly trained on labeled source-domain data and unlabeled target-domain data.

This paper instead targets the more general problem of source-free domain adaptation\textemdash also known as \emph{model adaptation}\textemdash for semantic segmentation.
In model adaptation, only (i) the model pre-trained on source images and (ii) unlabeled target images are available.
This pertains to many real-world use cases when the labeled source data is proprietary or inaccessible due to privacy concerns.
The complete absence of fine ground-truth annotations represents a significant challenge, as the model can easily drift and unlearn important concepts during adaptation.
To bolster the adaptation process, we leverage another form of weak supervision, which is far easier and cheaper to collect than pixel-wise semantic annotations.
In particular, multiple recent driving datasets\textemdash such as RobotCar~\cite{maddern20171}, ACDC~\cite{sakaridis2021acdc} and Boreas~\cite{burnett2022boreas}\textemdash traverse the same route several times under varying weather conditions, and provide GNSS-matched frames.
Each adverse-condition target image can thus be paired with a corresponding \emph{reference} image depicting roughly the same scene under normal conditions.
While also unlabeled, the reference images bridge the domain gap between the source and target domain by overlapping both with the source domain in terms of visual condition and with the target domain in terms of geography and sensor characteristics.

Our proposed method, named Contrastive Model Adaptation (CMA), leverages the reference predictions through a unified embedding space.
Assuming the reference and target images are sufficiently aligned,  co-located features should be similar between the two\textemdash neglecting dynamic objects and slight shifts in static content (\eg, missing leaves on a tree).
Accordingly, we posit that for a given target feature, its reference feature at the same spatial location should be \emph{closer in the embedding space than most other target features}.
An embedding space fulfilling this assumption would effectively eliminate condition-specific information, but simultaneously preserve semantic content.
We aim to create such an embedding space through contrastive learning, where dense spatial embeddings of the target image serve as \emph{anchors} (black patch in Fig.~\ref{fig:teaser}).
Each anchor is pulled towards a single \emph{positive}, \ie, the embedding of the reference image corresponding to the same location (green patch in Fig.~\ref{fig:teaser}).
Since the pre-trained source model is expected to produce semantic features of higher quality on the reference images than on the target images (for a qualitative comparison see appendix Sec.~\ref{sec:supp_source_preds}), this clustering step helps to correct less reliable anchor semantics.
Conversely, the anchor is pushed apart from the \emph{negatives}, which are simply target embeddings at other spatial locations (or from other target images, red patches in Fig.~\ref{fig:teaser}), to counteract mode collapse.
Through spatial alignment of the reference and target images and custom, confidence-modulated feature aggregation, we create robust embeddings for optimization with our cross-domain contrastive loss.

CMA yields state-of-the-art results for model adaptation on several normal-to-adverse semantic segmentation benchmarks. It even outperforms recent standard UDA methods on these benchmarks, despite its data handicap compared to the latter methods.
Attesting to our successful cross-domain embedding alignment, CMA delivers exceptionally \emph{robust} results, as shown by evaluations on the newly compiled Adverse-Condition Generalization (ACG) benchmark.
\section{Related Work}

\PAR{Model adaptation} or source-free domain adaptation lifts the assumption of standard unsupervised domain adaptation that data from the source domain are accessible at adaptation time, which renders the former task more challenging. In the absence of labeled data for providing supervision, model adaptation methods for semantic segmentation typically rely on loss-based constraints on the features and/or outputs of the network, which are computed for target images. Such losses promote robustness of the network to missing features~\cite{fleuret2021uncertainty,luo2022multilevel} or to perturbations of the inputs and features~\cite{luo2022multilevel}, or aim at minimizing entropy in the network outputs~\cite{fleuret2021uncertainty,wang2021tent}. Some works focus primarily on the normalization layers of the involved networks, encouraging consistency of the statistics of these layers across the initial source-trained model and the final adapted model~\cite{liu2021source} and optimizing channel-level affine transformations of the normalized features with respect to output entropy~\cite{wang2021tent}. 
Best ``upstream'' practices for training source models for model adaptation are explored in~\cite{kundu2021generalize}.
One previous work on model adaptation in semantic segmentation~\cite{huang2021model} has considered contrastive learning similar to ours. 
However, that work contrasts features within individual images across the model adaptation cycle, whereas we contrast features across domains due to multiple corresponding views.

\begin{figure*}
\centering
    \includegraphics[width=\linewidth]{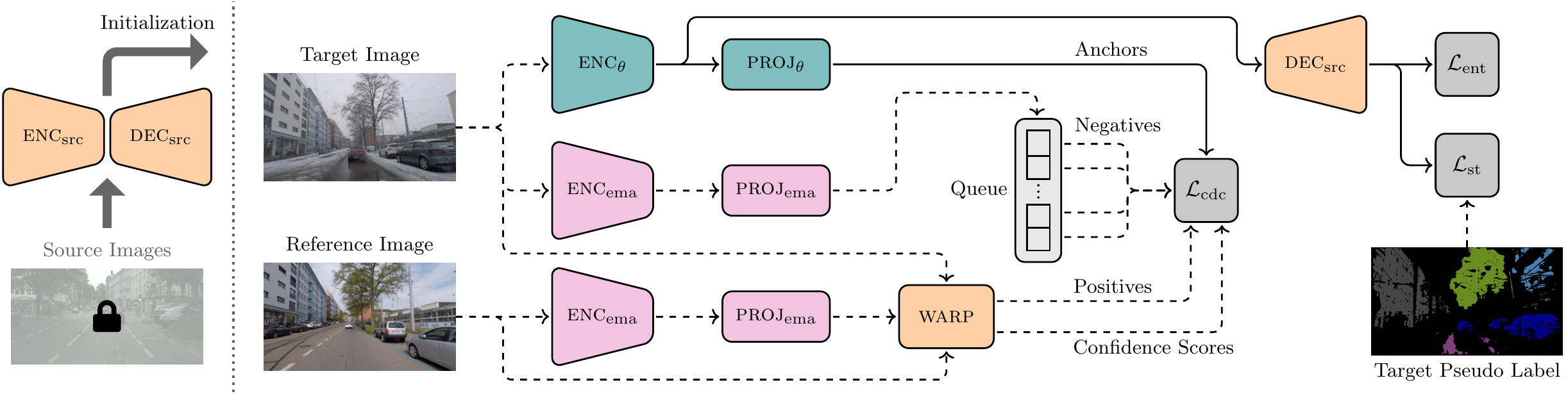}
    \caption{Overview of the CMA architecture. (Left) The segmentation network (\textnormal{\textsc{enc}} and \textnormal{\textsc{dec}}) is initialized with weights pre-trained on the source domain, however, access to the source data itself is prohibited. (Right) The model is trained with pairs of target and reference images. In addition to standard entropy minimization ($\mathcal{L}_\text{ent}$) and self-training ($\mathcal{L}_\text{st}$), we propose a cross-domain contrastive (CDC) loss ($\mathcal{L}_\text{cdc}$) to align features across domains. Dense embeddings are extracted from both images through projection heads \textnormal{\textsc{proj}}. The CDC loss pulls the target anchors close to corresponding reference embeddings (positives), while pushing them apart from other target embeddings stored in a queue (negatives). Crucially, the positives are obtained through spatial alignment (\textnormal{\textsc{warp}}) and robust feature aggregation (see Sec.~\ref{subsec:contr_feat_align}). Frozen modules are in orange, trainable modules in blue, and exponential moving average modules in pink. Gradients are only backpropagated through solid arrows.}
    \label{fig:method_overview}
\end{figure*}

\PAR{Contrastive learning} is a fundamental unsupervised framework based on instance discrimination for extracting informative representations. 
Seminal works on contrastive learning include \cite{oord2018representation}, which introduced the widely used InfoNCE loss, and \cite{chen2020simple}, which proposed a simple framework for visual contrastive learning.
While such fundamental works focus on the setting of unsupervised pre-training for image classification, there has been a body of recent literature examining contrastive learning for domain adaptive semantic segmentation, by primarily leveraging \emph{class-wise} contrast. 
\cite{jiang2022prototypical,lee2022bidirectional,xie2023sepico} employ partially dense contrast between classes using pixel features as anchors and class-level prototype vectors as positives and negatives. Along similar lines, \cite{li2021semantic,xie2023sepico} implement partially dense contrast between classes using pixel features as anchors and estimated class-level \emph{distributions} as positives and negatives, while \cite{marsden2022contrastive} use class prototypes both as anchors and as positives/negatives. These approaches are prone to false positive/negative samples which contaminate the contrastive loss due to potential errors in the target-domain pseudo-labels, which are used both to determine the anchors and to compute the class prototypes that serve as positives and negatives. By contrast, we propose a novel \emph{domain-wise} contrast, which does not rely on class (pseudo-)labels to construct the contrastive loss and avoids the above issue. Moreover, our cross-domain contrastive loss is \emph{fully} dense, in the sense that both the anchors and the positives/negatives are densely extracted in patches from the input images. In that regard, our method is closely related to~\cite{wang2021dense}, which also uses correspondences between matching views to contrast dense, locally aggregated features\textemdash albeit for unsupervised pre-training. Whereas correspondences between views are heuristically determined in~\cite{wang2021dense} through the similarities of backbone features, we have a dedicated, externally pre-trained module for computing correspondences via dense matching. 
Besides, contrary to~\cite{wang2021dense}, we explicitly combat false positive pairs owing to errors in the correspondences by incorporating 
confidence-guided feature aggregation: when grouping features locally, we weigh them by the confidence associated with the respective correspondences. In addition, low-confidence anchors and positives are filtered out.

\PAR{Cross-condition image-level correspondences} are provided by several driving datasets~\cite{maddern20171,sakaridis2020map,sakaridis2021acdc,burnett2022boreas} and can be utilized for normal-to-adverse condition domain adaptive semantic segmentation: 
\cite{larsson2019cross} find sparse, pixel-level correspondences and apply a consistency loss on semantic predictions.
Other works establish dense correspondences through two-view geometry~\cite{sakaridis2020map} or end-to-end dense matching~\cite{wu2021one,bruggemann2023refign}. The dense correspondences are subsequently used to enforce prediction consistency~\cite{wu2021one} or to fuse cross-condition predictions~\cite{sakaridis2020map,bruggemann2023refign}.
While we also use end-to-end dense matching to find correspondences, we instead use contrastive learning to create a condition-invariant, discriminative embedding space.
\section{Contrastive Model Adaptation (CMA)}

The aim of CMA is to fine-tune a given pre-trained semantic segmentation model on unlabeled, adverse-condition images. 
In contrast to standard unsupervised domain adaptation, access to the labeled source data\textemdash with which the model was originally trained\textemdash is assumed to be prohibited, \eg, due to privacy concerns. 
In the experimental setup for CMA, we are specifically given (i) a semantic segmentation model, pre-trained on a source dataset recorded under normal conditions, \eg, Cityscapes~\cite{cordts2016cityscapes}, and (ii) a set of unlabeled adverse-condition target samples, \eg, from ACDC~\cite{sakaridis2021acdc}, to whose population we aim to adapt the model.
Moreover, (iii) an unlabeled reference image is available for each target image, which depicts approximately the same scene as the adverse-condition target image, albeit under normal visual conditions.

\subsection{Architecture Overview}

Fig.~\ref{fig:method_overview} shows our model adaptation architecture.
The pre-trained source-model weights are used to initialize the encoder \textnormal{\textsc{enc}} and decoder \textnormal{\textsc{dec}}.
Since CMA focuses on generating condition-invariant, discriminative encoder features, decoder weights are kept frozen to preserve source-domain knowledge.
The encoder \textnormal{\textsc{enc}}$_\theta$ is adapted by three loss functions: entropy minimization ($\mathcal{L}_\text{ent}$, Sec.~\ref{subsec:total_loss_function}), self-training ($\mathcal{L}_\text{st}$, Sec.~\ref{subsec:total_loss_function}), and the proposed cross-domain contrastive (CDC) loss ($\mathcal{L}_\text{cdc}$, Sec.~\ref{subsec:contr_feat_align}).
We describe the individual modules in detail in the next sections.

\subsection{Spatial Alignment}
\label{subsec:alignment}

Although the reference-target image pairs depict the same scene, their viewpoint can differ substantially as they are only GNSS-matched.
The resulting correspondence discrepancies can have a detrimental effect when working on pixel-accurate tasks such as semantic segmentation.
We therefore densely warp the reference image into the viewpoint of the target image to obtain more accurate matches.

In particular, we choose the existing dense matching network UAWarpC~\cite{bruggemann2023refign} (\textnormal{\textsc{warp}} module in Fig.~\ref{fig:method_overview}), since it provides a confidence score for each displaced pixel, which is an important component of our downstream CDC loss (see Sec.~\ref{subsec:contr_feat_align}).
UAWarpC was independently trained in a self-supervised way on MegaDepth~\cite{li2018megadepth}, a large-scale collection of multi-view internet photos.
When training CMA, UAWarpC is frozen and warps the reference image features.
We tried to either warp the image before feeding it into the encoder \textnormal{\textsc{enc}}$_\theta$ or warp the dense feature maps and found that the latter works better.

\subsection{Cross-Domain Contrastive Loss}
\label{subsec:contr_feat_align}

The cross-domain contrastive (CDC) loss is the central component of CMA.
It incentivizes the encoder to learn features that discriminatively reflect the semantics, but are invariant to the visual condition.
To this end, spatial target image features represent \emph{anchors}, which are pulled towards spatially corresponding reference image features\textemdash the \emph{positives}. The anchors and positives are assumed to represent similar semantics in distinct visual conditions. Simultaneously, the anchors are contrasted to other target image features\textemdash the \emph{negatives}\textemdash to prevent mode collapse.
Although the CDC loss could in principle be applied to encoder features directly, we first project the dense features to a dedicated 128-dimensional embedding space, as per standard practice~\cite{chen2020simple}.
The projection head \textnormal{\textsc{proj}} consists of two 1\texttimes 1 convolutions with a ReLU non-linearity in-between.
As shown in Fig.~\ref{fig:method_overview}, the embeddings of the trainable model $\textnormal{\textsc{proj}}_\theta \circ \textnormal{\textsc{enc}}_\theta$ serve as anchors, while\textemdash as proposed in~\cite{he2020momentum}\textemdash positives and negatives are obtained by an exponential moving average model $\textnormal{\textsc{proj}}_\text{ema} \circ \textnormal{\textsc{enc}}_\text{ema}$ to improve their consistency.
Furthermore, we use a \emph{queue} to accumulate negatives~\cite{he2020momentum}.
This enables the use of a large number of negatives during instance discrimination, which encourages the learning of meaningful representations by making the discrimination more challenging.
Finally, the positives are spatially warped to align them with the anchors, as detailed in Sec.~\ref{subsec:alignment}.

Despite the warping, anchors and positives might not always depict the same semantic content, because (i) the warping described in Sec.~\ref{subsec:alignment} is not exact, \eg, street poles and other small or thin objects are often not perfectly aligned, and (ii) dynamic objects such as cars and pedestrians differ between the reference and target images.
Such false positives introduce excessive noise into the contrastive loss, which worsens generalization~\cite{tian2020makes}. 
To mitigate this issue, we use two strategies: patch-level grouping and confidence modulation.

\noindent \textbf{Patch-Level Grouping.} 
Inspired by~\cite{wang2021dense}, we average-pool spatial embeddings across square patches for anchors, positives, and negatives. 
However, differently from~\cite{wang2021dense}, we directly pool the embeddings, instead of applying pooling earlier in the model.
Due to the averaging, larger patches are more forgiving towards small errors in the warping, as well as small semantic discrepancies due to dynamic objects.
On the other hand, very large patch sizes do not promote the learning of local discriminative features.
The grid size dictating the employed grouping is thus a key hyperparameter; we choose a 7\texttimes 7 grid for square full-height crops of street images.
A desirable side-effect of patch-level grouping is a significant reduction in memory and computational cost.

\noindent \textbf{Confidence Modulation.} The confidence scores provided by the \textnormal{\textsc{warp}} module can be leveraged to refine and filter patch embeddings for anchors and positives.
We propose to use \emph{weighted} average pooling to create patch embeddings, where each pixel is weighted by its confidence score.
Accordingly, low-confidence correspondences within a single patch (\eg, resulting from pixels of a dynamic object) contribute less to the aggregated patch embedding.
In addition, patches with an average confidence of below 0.2 are deemed false positives and discarded altogether.

To formalize those steps, we define the set $\mathcal{N}_i$ to comprise all indices of pixels in the pooling receptive field of patch $i$.
$\mathbf{z}^a_j, \mathbf{z}^p_j \in \mathbb{R}^{128}$ are the \textnormal{\textsc{proj}} head outputs at pixel index $j$ for anchor and (warped) positive respectively.
$c_j \in [0, 1]$ is the corresponding warp confidence score (which is identical for anchor and positive). 
Importantly, $c_j=0$ for pixels without a valid correspondence.
Unnormalized patch embeddings for anchors $\mathbf{\tilde{a}}_i$ and positives $\mathbf{\tilde{p}}_i$ are computed through the weighted sums
\begin{equation}
    \mathbf{\tilde{a}}_i = \sum_{j \in \mathcal{N}_i} c_j \mathbf{z}^a_j,\quad \mathbf{\tilde{p}}_i = \sum_{j \in \mathcal{N}_i} c_j \mathbf{z}^p_j. 
\end{equation}
The embeddings are subsequently L2-normalized to obtain $\mathbf{a}_i$ and $\mathbf{p}_i$.
Meanwhile, negative embeddings $\mathbf{n}_j$ are obtained through simple average pooling, followed by L2-normalization.

To create an embedding space where, for each patch $i$, the anchor $\mathbf{a}_i$ is pulled towards the positive $\mathbf{p}_i$ and pushed away from $M$ negatives $\mathbf{n}_j$ (sourced from the queue of length $M$), we use the InfoNCE~\cite{oord2018representation} loss:
\begin{equation}
    \mathcal{L}_{\text{cdc},i} = -\log \frac{\exp{\left(\mathbf{a}_i^T \mathbf{p}_i / \tau \right)}}{\exp{\left(\mathbf{a}_i^T \mathbf{p}_i / \tau \right)} + \sum_{j=1}^M \exp{\left(\mathbf{a}_i^T \mathbf{n}_j / \tau \right)}} .
\end{equation}
$\tau$ is a temperature hyperparameter that scales the sensitivity of the loss function.
Finally, when aggregating the patch-wise losses, low-confidence patches are discarded:
\begin{equation}
    \mathcal{L}_{\text{cdc}} = \frac{\sum_{i} \mathcal{L}_{\text{cdc},i} [\bar{c}_i \geq 0.2]}{\sum_{i} [\bar{c}_i \geq 0.2]},
    \label{eq:discarding}
\end{equation}
where $[\cdot]$ denotes the Iverson bracket and $\bar{c}_i$ is the average-pooled confidence of patch $i$:
\begin{equation}
    \bar{c}_i = \frac{1}{|\mathcal{N}_i|} \sum_{j \in \mathcal{N}_i} c_j .
\end{equation}

\begin{figure}
    \adjustbox{max width=\linewidth}{%
        \input{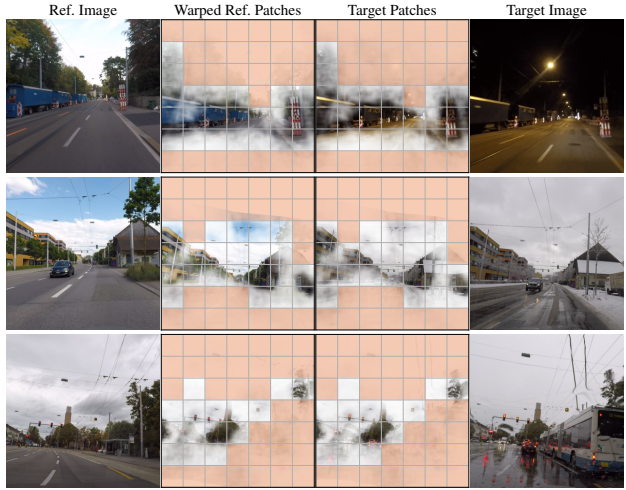}%
    }%
    \caption{Visualization of matched patches. The second column shows the warped reference image, with low-confident regions whited out. The drawn grid illustrates the patch-level grouping. Within each patch, features are aggregated proportionally to their confidence. Patches with low average confidence are discarded altogether, as shown by orange shading.}
    \label{fig:intuit}
    \vspace{-0.6em}
\end{figure}

The effect of patch-level grouping and confidence modulation is illustrated in Fig.~\ref{fig:intuit}.
In the two center columns, each pixel is whited out according to its confidence. Within each patch, the ``confident'', visible pixels are subsequently aggregated.
Orange patches are eliminated due to their overall low confidence.
Notice that the remaining features correspond well between the two images, despite the initial differences in viewpoint, dynamic objects, occlusions, \etc.
In fact, the warping confidence is rather conservative.

\subsection{Complete Training Loss}
\label{subsec:total_loss_function}

Besides the proposed CDC loss (Sec.~\ref{subsec:contr_feat_align}) we employ two commonly used loss functions.

\PAR{Self-Training.}
We follow the pseudo-labeling strategy of CBST~\cite{zou2018unsupervised} to create class-balanced pseudo-labels from confident predictions.
The pseudo-labels are created once before training by the source model, to inject regularization to the source.
We retain the most confident 20\% of pixels and all other pixels are ignored.
During model adaptation, we use a cross-entropy loss $\mathcal{L}_\text{st}$ for self-training.

\PAR{Entropy Minimization.}
We use entropy minimization as a regularizer during training.
$\mathcal{L}_\text{ent}$ is the mean normalized entropy of the predicted class probabilities over all pixels.

Finally, the complete training loss consists of a weighted sum of the three losses:
$\mathcal{L}_\text{tot} = \mathcal{L}_\text{st} + \lambda_\text{ent} \mathcal{L}_\text{ent} + \lambda_\text{cdc} \mathcal{L}_\text{cdc}$.
$\lambda_\text{ent}$ and $\lambda_\text{cdc}$ are hyperparameters that determine the relative importance given to the individual losses.

\section{Experiments}

In this section, we present extensive experimental results, comparing CMA to state-of-the-art model adaptation (Sec.~\ref{subsec:sota_comp}) and standard unsupervised domain adaptation (Sec.~\ref{subsec:standard_uda}) methods.
Moreover, we analyze generalization performance (Sec.~\ref{subsec:generalization}) and ablate various components (Sec.~\ref{subsec:ablations}) of the method.

\subsection{Setup}

\begin{table*}
\caption{Comparison to the state of the art in model adaptation on Cityscapes$\to$ACDC, with reported performance on the ACDC test set.}%
\smallskip%
\centering%
\resizebox{\linewidth}{!}{%
\ra{1.3}%
\begin{tabular}{@{}lccccccccccccccccccccccr@{}}\toprule
\multirow{2}{*}{Method} &&& \multicolumn{21}{c}{ACDC IoU\,$\uparrow$} \\
\cmidrule{4-24} &&& \rotatebox[origin=c]{90}{road} & \rotatebox[origin=c]{90}{sidew.} & \rotatebox[origin=c]{90}{build.} & \rotatebox[origin=c]{90}{wall} & \rotatebox[origin=c]{90}{fence} & \rotatebox[origin=c]{90}{pole} & \rotatebox[origin=c]{90}{light} & \rotatebox[origin=c]{90}{sign} & \rotatebox[origin=c]{90}{veget.} & \rotatebox[origin=c]{90}{terrain} & \rotatebox[origin=c]{90}{sky} & \rotatebox[origin=c]{90}{person} & \rotatebox[origin=c]{90}{rider} & \rotatebox[origin=c]{90}{car} & \rotatebox[origin=c]{90}{truck} & \rotatebox[origin=c]{90}{bus} & \rotatebox[origin=c]{90}{train} & \rotatebox[origin=c]{90}{motorc.} & \rotatebox[origin=c]{90}{bicycle} && \multicolumn{1}{c}{\phantom{00}\rotatebox[origin=c]{90}{\textbf{mean}}} \\ \midrule
Source model & \multirow{6}{*}{\rotatebox[origin=c]{90}{DeepLabv2~\cite{chen2017deeplab}}} && 76.6 & 40.5 & 56.0 & 12.0 & 27.3 & 35.6 & 40.2 & 45.6 & 69.8 & 38.2 & 76.2 & 21.3 & 12.4 & 65.6 & 25.2 & 29.2 & 28.1 & 15.2 & 34.6 && 39.5 \\
TENT~\cite{wang2021tent} &&& 84.1 & 48.0 & 56.9 & 21.1 & 30.3 & 43.7 & 56.3 & 53.7 & 69.6 & 36.7 & 61.8 & 55.0 & 33.0 & 78.7 & 40.6 & 43.0 & 48.7 & 30.5 & 39.3 && 49.0 \\
HCL~\cite{huang2021model} &&& 80.5 & 42.9 & 57.6 & 14.7 & 29.4 & 40.3 & 49.0 & 51.1 & 72.4 & 35.6 & 78.3 & 39.7 & 31.8 & 76.0 & 35.4 & 42.7 & 42.5 & 25.7 & 43.0 && 46.8 \\
URMA~\cite{fleuret2021uncertainty} &&& 85.4 & 52.9 & 62.9 & 20.4 & 34.4 & 39.9 & 36.7 & 43.9 & 74.9 & 46.9 & 85.1 & 27.2 & 22.4 & 76.0 & 40.5 & 41.5 & 38.9 & 20.6 & 46.2 && 47.2 \\
URMA + SimT~\cite{guo2022simt} &&& 83.5 & 52.7 & 60.7 & 19.6 & 33.7 & 42.0 & 43.1 & 47.4 & 75.0 & 42.5 & 85.8 & 39.8 & 19.6 & 76.9 & 39.6 & 42.7 & 41.1 & 24.0 & 43.1 && 48.0 \\
CMA &&& 83.1 & 52.7 & 65.4 & 18.7 & 30.5 & 44.5 & 56.3 & 53.9 & 76.7 & 39.7 & 79.0 & 54.2 & 31.2 & 76.7 & 40.2 & 39.3 & 47.4 & 29.8 & 38.6 && 50.4 \\
\midrule  
Source model & \multirow{6}{*}{\rotatebox[origin=c]{90}{SegFormer~\cite{xie2021segformer}}} && 85.7 & 51.0 & 76.6 & 36.4 & 37.1 & 45.2 & 55.7 & 57.5 & 77.7 & 52.0 & 84.1 & 60.3 & 34.8 & 82.9 & 61.6 & 65.4 & 73.4 & 37.9 & 52.5 && 59.4 \\
TENT~\cite{wang2021tent} &&& 84.0 & 51.5 & 75.4 & 36.8 & 37.2 & 46.2 & 56.0 & 57.7 & 77.9 & 52.9 & 81.7 & 59.9 & 36.0 & 82.9 & 60.8 & 65.5 & 73.6 & 38.3 & 52.4 && 59.3  \\
HCL~\cite{huang2021model} &&& 86.4 & 53.5 & 78.5 & 38.8 & 38.1 & 48.0 & 57.8 & 58.9 & 78.1 & 52.4 & 85.1 & 61.7 & 37.1 & 83.7 & 64.1 & 66.6 & 74.5 & 39.1 & 53.3 && 60.8 \\
URMA~\cite{fleuret2021uncertainty} &&& 89.2 & 60.4 & 84.3 & 48.7 & 42.5 & 53.8 & 65.4 & 63.8 & 76.3 & 57.3 & 85.9 & 63.4 & 43.9 & 85.8 & \textbf{68.8} & 73.2 & 82.8 & 46.3 & 48.4 && 65.3 \\
URMA + SimT~\cite{guo2022simt} &&& 90.0 & 65.7 & 80.6 & 46.0 & 41.7 & \textbf{56.3} & 65.2 & 62.7 & 75.9 & 55.6 & 84.4 & 66.4 & \textbf{46.6} & 85.4 & 68.4 & 72.3 & 80.0 & \textbf{46.8} & 58.0 && 65.7 \\
CMA &&& \textbf{94.0} & \textbf{75.2} & \textbf{88.6} & \textbf{50.5} & \textbf{45.5} & 54.9 & \textbf{65.7} & \textbf{64.2} & \textbf{87.1} & \textbf{61.3} & \textbf{95.2} & \textbf{67.0} & 45.2 & \textbf{86.2} & 68.6 & \textbf{76.6} & \textbf{83.9} & 43.3 & \textbf{60.5} && \textbf{69.1} \\
\bottomrule
\end{tabular}}%
\label{tab:acdc_sota}
\end{table*}

\begin{figure*}
    \adjustbox{max width=\textwidth}{%
        \input{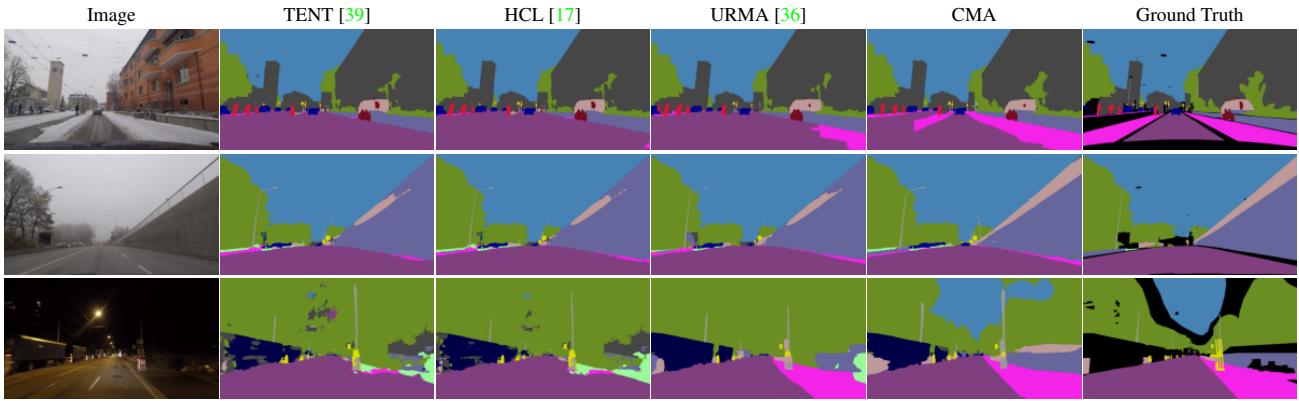}%
    }%
    \vspace{-1mm}
    \caption{Qualitative segmentation results of SegFormer-based model adaptation methods on ACDC validation images.}
    \label{fig:qualitative_seg}
\end{figure*}

\PAR{Datasets.} 
We use Cityscapes~\cite{cordts2016cityscapes} as a source dataset, and various target datasets: 
ACDC~\cite{sakaridis2021acdc} (train/val/test: 1600/406/2000 images), Dark Zurich~\cite{sakaridis2020map} (train/val/test: 2416/50/151 images), RobotCar Correspondence~\cite{maddern20171,larsson2019cross} (train/val/test:  6511/27/27 images), and CMU Correspondence~\cite{badino2011visual,larsson2019cross} (train/val/test: 28766/25/33 images).
All target datasets contain corresponding pairs of normal- and adverse-condition street images in the training set.
Adverse conditions vary between datasets: ACDC contains images in fog, night, rain, and snow; Dark Zurich consists of night images; RobotCar and CMU contain variable conditions as well as seasonal changes.
Unless otherwise stated, we report performance on the test sets for all datasets.

\PAR{ACG Benchmark.}
To assess the generalization performance of trained adverse-condition models, we present an adverse-condition generalization (ACG) benchmark consisting of diverse samples from several public street-scene segmentation datasets.
We inspected all labeled images of WildDash2~\cite{zendel2022unifying}, BDD100K~\cite{yu2020bdd100k}, Foggy Zurich~\cite{sakaridis2018model,dai2020curriculum}, and Foggy Driving~\cite{sakaridis2018semantic}, selected adverse-condition samples (featuring fog, night, rain, snow, or a combination of those), and manually verified the quality of each corresponding ground truth.
Samples with evident ground-truth inaccuracies were eliminated.
For Foggy Zurich and Foggy Driving, we also meticulously cross-checked every image for overlap with the ACDC dataset, as all three datasets were recorded in the same region.
We refer to appendix Sec.~\ref{sec:supp_acg_benchmark} for more details about the sample selection process and the resulting dataset statistics.
ACG consists of a highly \emph{diverse} set of 922 adverse-condition driving images from various geographical regions in Europe and North America.
We additionally divide ACG into 121 fog, 225 rain, 276 snow, and 300 night images, to allow for condition-wise evaluation.
Importantly, ACG-night also includes adverse-weather images, \eg, snowy nighttime scenes.
The curated list of ACG image filenames is publicly available via \url{https://github.com/brdav/cma}.

\PAR{Architectures and Hyperparameters.}
We conduct the bulk of our experiments using the state-of-the-art SegFormer~\cite{xie2021segformer} architecture, however, we also include experiments using DeepLabv2~\cite{chen2017deeplab}.
For all datasets and architectures, we train CMA for 10k iterations, where for the first 2.5k iterations we stop gradient flow from \textnormal{\textsc{proj}}$_\theta$ back to \textnormal{\textsc{enc}}$_\theta$ to ``warm up'' \textnormal{\textsc{proj}}$_\theta$. Since \textnormal{\textsc{proj}}$_\theta$ is the only randomly initialized module, we use a 10\texttimes\ learning rate for it w.r.t.\ \textnormal{\textsc{enc}}$_\theta$.
We find that a high momentum of 0.9999 works best for the exponential moving average components, presumably preserving source knowledge better during the adaptation process.
All models were trained on a single TITAN RTX GPU.
More details on training configurations are in appendix Sec.~\ref{sec:supp_training_details}.

\subsection{Comparison to Model Adaptation Methods}
\label{subsec:sota_comp}

We benchmark CMA against the state-of-the-art model adaptation methods TENT~\cite{wang2021tent}, HCL~\cite{huang2021model}, URMA~\cite{fleuret2021uncertainty}, and SimT~\cite{guo2022simt} on Cityscapes\textrightarrow ACDC and report the ACDC test set scores in Table~\ref{tab:acdc_sota}.
In this comparison, CMA is the only method using reference images.
Using a SegFormer architecture, CMA sets the new state of the art for model adaptation from Cityscapes to ACDC with a mIoU of 69.1\%, substantially outperforming all other methods.
Note that CMA outperforms competing methods both on static and dynamic classes, even though our contrastive loss does not explicitly target dynamic objects.
We hypothesize that the universal performance gain is enabled through the learned invariance to global condition-level variations, \eg, with respect to changes in scene illumination or reflectance.
For DeepLabv2, CMA obtains 50.4\% mIoU, still outperforming other methods, although in this case, all compared methods bring a substantial improvement over the source model.
Due to the large absolute performance difference, we conduct the rest of the experiments in this paper on the SegFormer architecture.
We show qualitative segmentation results for the different SegFormer-based methods on three ACDC validation set images in Fig.~\ref{fig:qualitative_seg}.
CMA predicts higher-quality segmentation maps than other methods, \eg, on the snowy sidewalk in the top image, the fence in the middle image, or the wall on the right in the bottom image.

We further evaluate CMA on three other adaptation settings: Cityscapes\textrightarrow Dark Zurich, Cityscapes\textrightarrow RobotCar, and Cityscapes\textrightarrow CMU in Table~\ref{tab:darkzurich_robotcar_cmu_sota}.
For all investigated scenarios, CMA significantly outperforms other methods.

\begin{table}
\caption{Comparison to the state of the art in model adaptation on Cityscapes$\to$Dark Zurich, Cityscapes$\to$RobotCar, and Cityscapes$\to$CMU. All models use a SegFormer architecture.}%
\smallskip%
\centering%
\resizebox{\columnwidth}{!}{%
\ra{1.3}%
\setlength{\tabcolsep}{9pt}%
\begin{tabular}{@{}lcccr@{}}\toprule
\multirow{2}{*}{Method} && \multicolumn{3}{c}{mIoU\,$\uparrow$} \\
\cmidrule{3-5} && Dark Zurich~\cite{sakaridis2020map} & RobotCar~\cite{maddern20171,larsson2019cross} & CMU~\cite{badino2011visual,larsson2019cross}\\
\midrule
Source model & \multirow{6}{*}{\rotatebox[origin=c]{90}{SegFormer~\cite{xie2021segformer}}} & 41.7 & 50.0 & 80.0 \\
TENT~\cite{wang2021tent} && 42.8 & 50.1 & 78.9  \\
HCL~\cite{huang2021model} && 42.7 & 50.1 & 80.2  \\
URMA~\cite{fleuret2021uncertainty} && 49.3 & 51.6 & 82.8 \\
URMA + SimT~\cite{guo2022simt} && 50.1 & 52.4 & 83.9 \\
CMA && \textbf{53.6} & \textbf{54.3} & \textbf{92.0} \\
\bottomrule
\end{tabular}}%
\label{tab:darkzurich_robotcar_cmu_sota}
\end{table}

\subsection{Comparison to Standard UDA Methods}
\label{subsec:standard_uda}

Table~\ref{tab:comparison_to_uda} shows a comparison of CMA to standard unsupervised domain adaptation (UDA) methods, which use the labeled source data during the adaptation process.
Standard UDA methods are thus not susceptible to forgetting source knowledge.
Despite this handicap, CMA compares favorably to DAFormer~\cite{hoyer2022daformer}, SePiCo~\cite{xie2023sepico}, and HRDA~\cite{hoyer2022hrda} on Cityscapes\textrightarrow ACDC, while only slightly falling behind on the more challenging Cityscapes\textrightarrow Dark Zurich.

\begin{table}
\caption{Comparison of CMA to state-of-the-art standard UDA methods. Adaptation from Cityscapes as source dataset.}%
\smallskip%
\centering%
\resizebox{\linewidth}{!}{%
\ra{1.3}%
\setlength{\tabcolsep}{12pt}%
\begin{tabular}{@{}lcccr@{}}\toprule
\multirow{2}{*}{Method} & \multirow{2}{*}{Source-Free} && \multicolumn{2}{c}{mIoU\,$\uparrow$} \\
\cmidrule{4-5} 
&&& ACDC~\cite{sakaridis2021acdc} & Dark Zurich~\cite{sakaridis2020map} \\
\midrule
DAFormer~\cite{hoyer2022daformer} &&& 55.4 & 53.8 \\
SePiCo~\cite{xie2023sepico} &&& 59.1 & 54.2 \\
HRDA~\cite{hoyer2022hrda} &&& 68.0 & \textbf{55.9} \\
CMA & \checkmark && \textbf{69.1} & 53.6 \\
\bottomrule
\end{tabular}}%
\label{tab:comparison_to_uda}
\end{table}

\subsection{Generalization and Robustness}
\label{subsec:generalization}

To evaluate the generalization performance of trained adverse-condition models, we test the Cityscapes\textrightarrow ACDC models on our diverse ACG benchmark.
We report the condition-wise mIoU for fog, night, rain, and snow, as well as the overall mIoU, in Table~\ref{tab:generalization}.
CMA achieves the best generalization performance compared to other model adaptation and UDA methods, with a mIoU of 51.3\% for all ACG samples.
Interestingly, CMA performs exceptionally well on the most challenging ACG-night split, which also contains combinations of conditions (\eg, night and rain), which are absent from the training set.
This corroborates that CMA learns highly \emph{robust} representations through our proposed contrastive cross-domain feature alignment.

\begin{table}
\caption{ACG benchmark generalization performance of models adapted from Cityscapes to ACDC.}%
\smallskip%
\centering%
\resizebox{\linewidth}{!}{%
\ra{1.3}%
\setlength{\tabcolsep}{8pt}%
\begin{tabular}{@{}lccccccccr@{}}\toprule
\multirow{2}{*}{Method} & \multirow{2}{*}{Source-Free} && \multicolumn{5}{c}{ACG mIoU\,$\uparrow$}\\
\cmidrule{4-8} 
&&& fog & night & rain & snow & all \\
\midrule
TENT~\cite{wang2021tent} & \checkmark && 52.6 & 27.7 & 47.5 & 41.1 & 40.0 \\
HCL~\cite{huang2021model} & \checkmark && 54.2 & 28.3 & 48.2 & 42.4 & 40.8 \\
URMA~\cite{fleuret2021uncertainty} & \checkmark && 54.1 & 31.0 & 51.9 & 45.5 & 44.4 \\
DAFormer~\cite{hoyer2022daformer} &&& 52.6 & 21.5 & 47.5 & 33.6 & 40.1 \\
SePiCo~\cite{xie2023sepico} &&& 53.9 & 20.6 & 46.3 & 36.1 & 38.6 \\
HRDA~\cite{hoyer2022hrda} &&& \textbf{60.0} & 27.1 & 56.2 & 43.3 & 48.9 \\
CMA & \checkmark && 59.7 & \textbf{40.0} & \textbf{59.6} & \textbf{52.2} & \textbf{51.3} \\
\bottomrule
\end{tabular}}%
\label{tab:generalization}
\end{table}

\begin{table*}
\caption{Ablation study on the ACDC validation set, reporting IoU. ``EMA'': exponential moving average model for positives and negatives.}%
\smallskip%
\centering%
\resizebox{\linewidth}{!}{%
    \ra{1.3}%
%
%
\pgfplotsset{%
    colormap/RdBu-5,
}%
\pgfplotstableset{%
    header=false,
    col sep=&,
    row sep=\\,
    every head row/.style={output empty row},  
    every first row/.style={before row=\midrule},
    every last row/.style={after row=\bottomrule},
    /color cells/min/.initial=0,
    /color cells/max/.initial=1000,
    /color cells/textcolor/.initial=,
    %
    color cells/.code={%
        \pgfqkeys{/color cells}{#1}%
        \pgfkeysalso{%
            postproc cell content/.code={%
                \begingroup
                %
                \pgfkeysgetvalue{/pgfplots/table/@preprocessed cell content}\value
                \ifx\value\empty
                    \endgroup
                \else
                \pgfmathfloatparsenumber{\value}%
                \pgfmathfloattofixed{\pgfmathresult}%
                \let\value=\pgfmathresult
                %
                \pgfplotscolormapaccess
                    [\pgfkeysvalueof{/color cells/min}:\pgfkeysvalueof{/color cells/max}]
                    {\value}
                    {\pgfkeysvalueof{/pgfplots/colormap name}}%
                %
                \pgfkeysgetvalue{/pgfplots/table/@cell content}\typesetvalue
                \pgfkeysgetvalue{/color cells/textcolor}\textcolorvalue
                %
                \toks0=\expandafter{\typesetvalue}%
                \xdef\temp{%
                    \noexpand\pgfkeysalso{%
                        @cell content={%
                            \noexpand\cellcolor[rgb]{\pgfmathresult}%
                            \noexpand\definecolor{mapped color}{rgb}{\pgfmathresult}%
                            \ifx\textcolorvalue\empty
                            \else
                                \noexpand\color{\textcolorvalue}%
                            \fi
                            \the\toks0 %
                        }%
                    }%
                }%
                \endgroup
                \temp
                \fi
            }%
        }%
    }
}%
\def\mycr{13.7}
\pgfmathsetmacro{\roadmin}{90.1 - \mycr}%
\pgfmathsetmacro{\roadmax}{90.1 + \mycr}%
\pgfmathsetmacro{\sidewmin}{64.6 - \mycr}%
\pgfmathsetmacro{\sidewmax}{64.6 + \mycr}%
\pgfmathsetmacro{\buildmin}{79.0 - \mycr}%
\pgfmathsetmacro{\buildmax}{79.0 + \mycr}%
\pgfmathsetmacro{\wallmin}{39.0 - \mycr}%
\pgfmathsetmacro{\wallmax}{39.0 + \mycr}%
\pgfmathsetmacro{\fencemin}{33.4 - \mycr}%
\pgfmathsetmacro{\fencemax}{33.4 + \mycr}%
\pgfmathsetmacro{\polemin}{54.0 - \mycr}%
\pgfmathsetmacro{\polemax}{54.0 + \mycr}%
\pgfmathsetmacro{\lightmin}{73.6 - \mycr}%
\pgfmathsetmacro{\lightmax}{73.6 + \mycr}%
\pgfmathsetmacro{\signmin}{56.0 - \mycr}%
\pgfmathsetmacro{\signmax}{56.0 + \mycr}%
\pgfmathsetmacro{\vegetmin}{71.0 - \mycr}%
\pgfmathsetmacro{\vegetmax}{71.0 + \mycr}%
\pgfmathsetmacro{\terrainmin}{35.9 - \mycr}%
\pgfmathsetmacro{\terrainmax}{35.9 + \mycr}%
\pgfmathsetmacro{\skymin}{82.1 - \mycr}%
\pgfmathsetmacro{\skymax}{82.1 + \mycr}%
\pgfmathsetmacro{\personmin}{61.8 - \mycr}%
\pgfmathsetmacro{\personmax}{61.8 + \mycr}%
\pgfmathsetmacro{\ridermin}{37.2 - \mycr}%
\pgfmathsetmacro{\ridermax}{37.2 + \mycr}%
\pgfmathsetmacro{\carmin}{84.9 - \mycr}%
\pgfmathsetmacro{\carmax}{84.9 + \mycr}%
\pgfmathsetmacro{\truckmin}{71.9 - \mycr}%
\pgfmathsetmacro{\truckmax}{71.9 + \mycr}%
\pgfmathsetmacro{\busmin}{69.7 - \mycr}%
\pgfmathsetmacro{\busmax}{69.7 + \mycr}%
\pgfmathsetmacro{\trainmin}{55.1 - \mycr}%
\pgfmathsetmacro{\trainmax}{55.1 + \mycr}%
\pgfmathsetmacro{\motorcmin}{46.9 - \mycr}%
\pgfmathsetmacro{\motorcmax}{46.9 + \mycr}%
\pgfmathsetmacro{\bicyclemin}{35.7 - \mycr}%
\pgfmathsetmacro{\bicyclemax}{35.7 + \mycr}%
\pgfmathsetmacro{\meanmin}{60.1 - \mycr}%
\pgfmathsetmacro{\meanmax}{60.1 + \mycr}%
\pgfplotstableread[header=true]{
road & sidew. & build. & wall & fence & pole & light & sign & veget. & terrain & sky & person & rider & car & truck & bus & train & motorc. & bicycle & mean \\
90.1 & 64.6 & 79.0 & 39.0 & 33.4 & 54.0 & 73.6 & 56.0 & 71.0 & 35.9 & 82.1 & 61.8 & 37.2 & 84.9 & 71.9 & 69.7 & 55.1 & 46.9 & 35.7 & 60.1 \\ 
92.2 & 69.1 & 82.5 & 45.9 & 40.8 & 56.0 & 73.4 & 58.8 & 78.2 & 40.2 & 87.6 & 64.0 & 41.4 & 85.4 & 79.5 & 72.8 & 65.2 & 48.5 & 46.9 & 64.7 \\ 
92.3 & 68.9 & 82.8 & 44.2 & 39.0 & 54.1 & 73.1 & 58.2 & 81.6 & 39.8 & 91.0 & 63.8 & 41.9 & 83.0 & 74.0 & 72.5 & 64.9 & 45.0 & 37.8 & 63.6 \\ 
92.5 & 70.0 & 82.7 & 46.0 & 40.9 & 56.8 & 73.9 & 59.5 & 77.7 & 40.2 & 86.9 & 64.4 & 41.8 & 85.5 & 80.2 & 72.9 & 66.3 & 48.2 & 48.8 & 65.0 \\ 
93.3 & 72.0 & 84.7 & 47.4 & 41.2 & 57.8 & 75.1 & 60.7 & 83.1 & 42.8 & 92.4 & 64.4 & 40.4 & 84.7 & 77.9 & 73.9 & 64.7 & 48.8 & 47.7 & 65.9 \\ 
93.3 & 72.3 & 84.9 & 47.7 & 41.4 & 59.1 & 75.9 & 61.3 & 84.1 & 44.2 & 93.3 & 65.9 & 40.8 & 85.2 & 81.6 & 74.0 & 65.8 & 50.1 & 49.5 & 66.9 \\ 
94.7 & 75.6 & 85.4 & 48.0 & 43.3 & 59.4 & 75.9 & 61.3 & 84.8 & 44.3 & 93.6 & 65.8 & 39.6 & 85.7 & 81.6 & 73.8 & 69.0 & 48.8 & 47.1 & 67.2 \\ 
}\mytable%
\def\mycolw{\textnormal{\textsc{alignn}} }
\begin{tabular}{@{}c@{}c@{}c@{}c@{}c@{}c@{}c@{}c@{}c@{}c@{}c@{}c@{}c@{}c@{}c@{}c@{}c@{}c@{}c@{}c@{}c@{}c@{}c@{}c@{}c@{}r@{}}\toprule
& \textover[c]{$\mathcal{L}_\text{cdc}$}{\mycolw} & \textover[c]{\begin{tabular}{@{}c@{}}patch-level\\[-2pt] grouping\end{tabular}}{\mycolw} & \textover[c]{warp}{\mycolw} & \textover[c]{\begin{tabular}{@{}c@{}}confidence-\\[-2pt] modulation\end{tabular}}{\mycolw} & \textover[c]{EMA}{\mycolw} & \rotatebox[origin=c]{90}{road} & \rotatebox[origin=c]{90}{sidew.} & \rotatebox[origin=c]{90}{build.} & \rotatebox[origin=c]{90}{wall} & \rotatebox[origin=c]{90}{fence} & \rotatebox[origin=c]{90}{pole} & \rotatebox[origin=c]{90}{light} & \rotatebox[origin=c]{90}{sign} & \rotatebox[origin=c]{90}{veget.} & \rotatebox[origin=c]{90}{terrain} & \rotatebox[origin=c]{90}{sky} & \rotatebox[origin=c]{90}{person} & \rotatebox[origin=c]{90}{rider} & \rotatebox[origin=c]{90}{car} & \rotatebox[origin=c]{90}{truck} & \rotatebox[origin=c]{90}{bus} & \rotatebox[origin=c]{90}{train} & \rotatebox[origin=c]{90}{motorc.} & \rotatebox[origin=c]{90}{bicycle} & \multicolumn{1}{c}{\phantom{0}\rotatebox[origin=c]{90}{\textbf{mean}}} \\
{\pgfplotstabletypeset[string type]{
\textcolor{gray}{1} \\ 
\textcolor{gray}{2} \\ 
\textcolor{gray}{3} \\ 
\textcolor{gray}{4} \\ 
\textcolor{gray}{5} \\ 
\textcolor{gray}{6} \\ 
\textcolor{gray}{7} \\ 
}} &
{\pgfplotstabletypeset[string type]{
\textover[c]{\phantom{\checkmark}}{\mycolw} \\ 
\checkmark \\ 
\checkmark \\
\checkmark \\
\checkmark \\
\checkmark \\
\checkmark \\
}} &
{\pgfplotstabletypeset[string type]{
\textover[c]{\phantom{\checkmark}}{\mycolw} \\ 
\phantom{\checkmark} \\ 
\checkmark \\
\phantom{\checkmark} \\
\checkmark \\
\checkmark \\
\checkmark \\
}} &
{\pgfplotstabletypeset[string type]{
\textover[c]{\phantom{\checkmark}}{\mycolw} \\ 
\phantom{\checkmark} \\ 
\phantom{\checkmark} \\ 
\checkmark \\
\checkmark \\
\checkmark \\
\checkmark \\
}} &
{\pgfplotstabletypeset[string type]{
\textover[c]{\phantom{\checkmark}}{\mycolw} \\ 
\phantom{\checkmark} \\ 
\phantom{\checkmark} \\ 
\phantom{\checkmark} \\ 
\phantom{\checkmark} \\ 
\checkmark \\
\checkmark \\
}} &
{\pgfplotstabletypeset[string type]{
\textover[c]{\phantom{\checkmark}}{\mycolw} \\ 
\phantom{\checkmark} \\ 
\phantom{\checkmark} \\ 
\phantom{\checkmark} \\ 
\phantom{\checkmark} \\ 
\phantom{\checkmark} \\ 
\checkmark \\
}} &
{\pgfplotstabletypeset[
    columns={road},
    color cells={min=\roadmin,max=\roadmax},
    fixed zerofill,
    precision=1,
]\mytable} &
{\pgfplotstabletypeset[
    columns={sidew.},
    color cells={min=\sidewmin,max=\sidewmax},
    fixed zerofill,
    precision=1,
]\mytable} &
{\pgfplotstabletypeset[
    columns={build.},
    color cells={min=\buildmin,max=\buildmax},
    fixed zerofill,
    precision=1,
]\mytable} &
{\pgfplotstabletypeset[
    columns={wall},
    color cells={min=\wallmin,max=\wallmax},
    fixed zerofill,
    precision=1,
]\mytable} &
{\pgfplotstabletypeset[
    columns={fence},
    color cells={min=\fencemin,max=\fencemax},
    fixed zerofill,
    precision=1,
]\mytable} &
{\pgfplotstabletypeset[
    columns={pole},
    color cells={min=\polemin,max=\polemax},
    fixed zerofill,
    precision=1,
]\mytable} &
{\pgfplotstabletypeset[
    columns={light},
    color cells={min=\lightmin,max=\lightmax},
    fixed zerofill,
    precision=1,
]\mytable} &
{\pgfplotstabletypeset[
    columns={sign},
    color cells={min=\signmin,max=\signmax},
    fixed zerofill,
    precision=1,
]\mytable} &
{\pgfplotstabletypeset[
    columns={veget.},
    color cells={min=\vegetmin,max=\vegetmax},
    fixed zerofill,
    precision=1,
]\mytable} &
{\pgfplotstabletypeset[
    columns={terrain},
    color cells={min=\terrainmin,max=\terrainmax},
    fixed zerofill,
    precision=1,
]\mytable} &
{\pgfplotstabletypeset[
    columns={sky},
    color cells={min=\skymin,max=\skymax},
    fixed zerofill,
    precision=1,
]\mytable} &
{\pgfplotstabletypeset[
    columns={person},
    color cells={min=\personmin,max=\personmax},
    fixed zerofill,
    precision=1,
]\mytable} &
{\pgfplotstabletypeset[
    columns={rider},
    color cells={min=\ridermin,max=\ridermax},
    fixed zerofill,
    precision=1,
]\mytable} &
{\pgfplotstabletypeset[
    columns={car},
    color cells={min=\carmin,max=\carmax},
    fixed zerofill,
    precision=1,
]\mytable} &
{\pgfplotstabletypeset[
    columns={truck},
    color cells={min=\truckmin,max=\truckmax},
    fixed zerofill,
    precision=1,
]\mytable} &
{\pgfplotstabletypeset[
    columns={bus},
    color cells={min=\busmin,max=\busmax},
    fixed zerofill,
    precision=1,
]\mytable} &
{\pgfplotstabletypeset[
    columns={train},
    color cells={min=\trainmin,max=\trainmax},
    fixed zerofill,
    precision=1,
]\mytable} &
{\pgfplotstabletypeset[
    columns={motorc.},
    color cells={min=\motorcmin,max=\motorcmax},
    fixed zerofill,
    precision=1,
]\mytable} &
{\pgfplotstabletypeset[
    columns={bicycle},
    color cells={min=\bicyclemin,max=\bicyclemax},
    fixed zerofill,
    precision=1,
]\mytable} &
{\pgfplotstabletypeset[
    columns={mean},
    color cells={min=\meanmin,max=\meanmax},
    fixed zerofill,
    precision=1,
    columns/0/.style={column type=r},%
]\mytable}%
\end{tabular}
    }
\label{tab:ablations}
\end{table*}

\subsection{Ablation Study and Further Analysis}
\label{subsec:ablations}

All the numbers in this section are from the ACDC validation set.
We report the mean performance over 3 repeated runs for each experiment, to reduce variance.

\PAR{Ablation Study.} 
Table~\ref{tab:ablations} shows the ablation study for several important CMA components.
Row 1 represents CMA without the CDC loss, solely relying on entropy minimization and self-training for adaptation.
A comparison of row 1 to the final model in row 7 reveals the large performance increase of 7.1\% mIoU owing to the CDC loss.
In row 2 the contrastive loss is added, but the embeddings are obtained by global average pooling over the entire image.
Nevertheless, this improves performance substantially by 4.6\% mIoU.
Next, we add patch-level grouping on a 7\texttimes 7 grid combined with reference image warping in row 5, which brings a 1.2\% mIoU improvement over the global embeddings of row 2.
Interestingly, using patches without warping decreases the performance, as shown in row 3.
This can be explained by the excessive noise introduced to the contrastive loss due to patch misalignment between reference and target.
A comparison of rows 4 and 5 reveals that, although warping is responsible for the majority of the performance gain, patch-level grouping brings further improvements by producing more locally discriminative features.
In row 6, we show that adding our confidence modulation to patch forming leads to another 1\% mIoU increase.
Finally, row 7 refers to the complete model, which involves estimating positives and negatives through an exponential moving average model, instead of simply using the source model and a random projection head.

\PAR{Effect of Reference Images.}
Compared to other model adaptation methods, CMA uses extra reference images through the CDC loss. 
We, therefore, train two baseline models\textemdash CMA without the CDC loss and URMA\textemdash by including the reference images in two ways: 
(i) mixing the reference and target images randomly, and adapting to the combination of both, and
(ii) adapting in a curriculum, \ie, first adapting from the source to the reference domain, and then using the same method to continue adapting from the reference to the target domain.
Table~\ref{tab:curriculum_and_mix} shows that our contrastive approach clearly outperforms both of these simple strategies.
The performance benefits of our approach are thus not attainable by naively introducing reference images.

\begin{table}
\caption{Comparison of different strategies for involving reference images in model adaptation, reporting ACDC validation scores.}%
\smallskip%
\centering%
\resizebox{\columnwidth}{!}{%
\ra{1.3}%
\setlength{\tabcolsep}{14pt}%
\begin{tabular}{@{}lcccr@{}}\toprule
Model & Target & Reference & Type & mIoU\,$\uparrow$ \\
\midrule 
Source model~\cite{xie2021segformer} & & & - & 56.6 \\
\midrule 
URMA~\cite{fleuret2021uncertainty} & \checkmark & & - & 63.2 \\
URMA~\cite{fleuret2021uncertainty} & \checkmark & \checkmark & mixed & 62.9 \\
URMA~\cite{fleuret2021uncertainty} & \checkmark & \checkmark & curriculum & 64.1 \\
\midrule 
CMA w/o CDC loss & \checkmark & & - &  60.1 \\
CMA w/o CDC loss & \checkmark & \checkmark & mixed & 60.3  \\
CMA w/o CDC loss & \checkmark & \checkmark & curriculum & 61.5\\
\midrule 
CMA & \checkmark & \checkmark & contrastive & 67.2 \\
\bottomrule
\end{tabular}}%
\label{tab:curriculum_and_mix}
\end{table}

\PAR{Alternative Contrastive Loss.}
In addition to the mechanisms discussed in Sec.~\ref{subsec:contr_feat_align}, we explored swapping the InfoNCE loss for more robust alternatives, to reduce the detrimental effects of false positives or false negatives.
We picked the debiased contrastive loss of \cite{chuang2020debiased} to account for false negatives, and the robust InfoNCE (RINCE)~\cite{chuang2022robust} loss to account for false positives.
As shown in Table~\ref{tab:contrastive_losses}, neither alternative shows significant improvements over InfoNCE.
Even though RINCE performs slightly better overall, it introduces extra complexity, which prompts us to prefer the simpler InfoNCE loss.
The negligible benefit of RINCE implies that patch-level grouping and confidence modulation already effectively mitigate the false positive rate. 

\begin{table}
\caption{CMA with alternative contrastive loss functions.}%
\smallskip%
\centering%
\resizebox{\columnwidth}{!}{%
\ra{1.3}%
\setlength{\tabcolsep}{22pt}%
\begin{tabular}{@{}lcccr@{}}\toprule
&& Debiased~\cite{chuang2020debiased} & RINCE~\cite{chuang2022robust} & InfoNCE~\cite{oord2018representation} \\
\midrule 
CMA && 66.3 & 67.4 & 67.2 \\
\bottomrule
\end{tabular}}%
\label{tab:contrastive_losses}
\end{table}

\begin{figure}
    \adjustbox{max width=\linewidth}{%
        \input{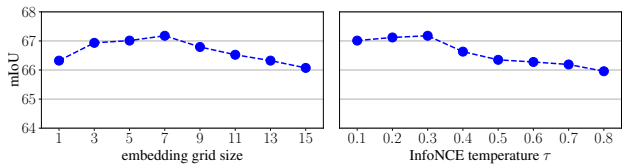}%
    }%
    \vspace{-0.2mm}%
    \caption{Hyperparameter study of the embedding grid size and the temperature in the InfoNCE loss.}
    \label{fig:grid_search}
\end{figure}

\PAR{Hyperparameter Sensitivity.} 
Fig.~\ref{fig:grid_search} shows the sensitivity of CMA performance to changes in two central, method-specific hyperparameters: the embedding grid size, and the InfoNCE temperature $\tau$.
Note that the performance is quite insensitive to either hyperparameter.

\begin{figure}
    \adjustbox{max width=\linewidth}{%
        \input{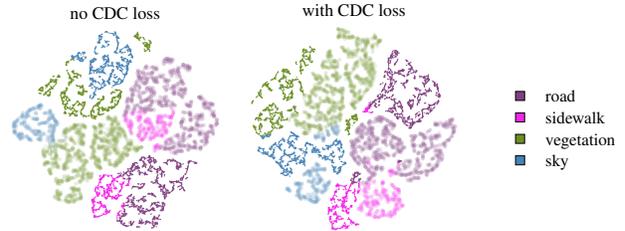}%
    }%
    \vspace{-0.2mm}%
    \caption{t-SNE plots showing semantic features of a pair of corresponding ACDC validation set images (adverse$\leftrightarrow$sharp, normal$\leftrightarrow$blurry), for a CMA model trained without (left) and with (right) the CDC loss. For clarity, only four classes are shown.}
    \label{fig:tsne}
\end{figure}

\PAR{Embedding Space Visualization.} 
The t-SNE~\cite{van2008visualizing} visualizations in Fig.~\ref{fig:tsne} show semantic features extracted from a corresponding pair of adverse- and normal-condition images of ACDC.
The features are color-coded by ground truth class, whereby adverse-condition features are plotted sharp and normal-condition features blurry (the ``ground truth'' for the normal-condition image was obtained through pseudo-labeling).
For clarity, only road, sidewalk, vegetation, and sky features are plotted.
The left plot shows the features of CMA without the CDC loss.
Note that sky (blue) and sidewalk (pink) features are scattered.
By contrast, with the CDC loss, the features of these classes are correctly grouped together across conditions in the right plot.

\section{Conclusion}

We present CMA, a model adaptation method for cross-condition semantic segmentation.
CMA leverages image-level correspondences to learn condition-invariant features through a contrastive loss.
This fosters a shared embedding space, where adverse-condition image features are clustered with semantically corresponding normal-condition features.
As experimentally shown, this leads to large performance gains in normal-to-adverse model adaptation, with CMA setting the new state of the art on several benchmarks.

\PAR{Acknowledgment.} This work was supported by the ETH Future Computing Laboratory (EFCL), financed by a donation from Huawei Technologies.

{\small
\bibliographystyle{ieee_fullname}
\bibliography{references}
}

\begin{appendices}

\beginappendixa
\section{Training Details}
\label{sec:supp_training_details}

We provide additional training details in this section. All models were trained using Automatic Mixed Precision on a single consumer TITAN RTX GPU.

\subsection{Optimization}
\label{subsec:supp_optim}

We train for 10k iterations for all datasets.
During the first 2.5k iterations, gradient backpropagation from the projection head to the backbone is stopped, to avoid noisy weight updates on the pretrained backbone weights.
We use an AdamW~\cite{loshchilov2017decoupled} optimizer with weight decay 0.01 and a linear learning rate decay with linear warm-up for the first 1500 iterations.
The chosen learning rates for CMA are $1\times 10^{-5}$ (SegFormer-based) and $2\times 10^{-8}$ (DeepLabv2-based), using 1 (SegFormer-based) or 2 (DeepLabv2-based) adverse-reference image pairs per batch.
For the weights of the projection head, the learning rate is multiplied by a factor of 10, as it is initialized randomly.
The individual loss weights are $\lambda_\text{ent} = 0.01$ and $\lambda_\text{cdc} = 1.0$ for SegFormer-based CMA and $\lambda_\text{ent} = 1.0$ and $\lambda_\text{cdc} = 1.0$ for DeepLabv2-based CMA.

\subsection{CDC Loss Hyperparameters}
\label{subsec:cdc_loss_hyper}

For partitioning the dense feature map into patches, a 7\texttimes 7 grid is used.
Positives and negatives are encoded with an exponential moving average network using a momentum of 0.9999. 
Negatives are then stored in a queue of size 65536.
The temperature $\tau$ of the InfoNCE loss varies depending on the dataset, we use 0.3 for ACDC, 0.03 for Dark Zurich, 0.3 for RobotCar, and 0.1 for CMU.
For eliminating unreliable patches in the confidence modulation, a threshold of 0.2 is used throughout.

\subsection{Data Handling}

Training data augmentation consists of random cropping to square shape\textemdash such that the crop size coincides with the shorter sidelength of the input\textemdash and random horizontal flipping.
Note that no resizing is applied.

Test predictions are generated through a sliding window approach.
The windows are square, with a sidelength equal to the shorter input sidelength.
Consecutive windows overlap for between 0\% and 50\% of their sidelength, depending on the input aspect ratio.

\subsection{Baselines}

We reimplemented the baselines TENT~\cite{wang2021tent}, HCL~\cite{huang2021model}, and URMA~\cite{fleuret2021uncertainty} for a fair comparison, carefully following their published code for reference.
The learning rate, loss weights, as well as method-specific hyperparameters were separately tuned for each method.
For SegFormer-based HCL, we had to introduce random subsampling of anchors for the contrastive loss, due to prohibitive memory demands otherwise.

\beginappendixb

\begin{figure*}
    \adjustbox{max width=\textwidth}{%
        \input{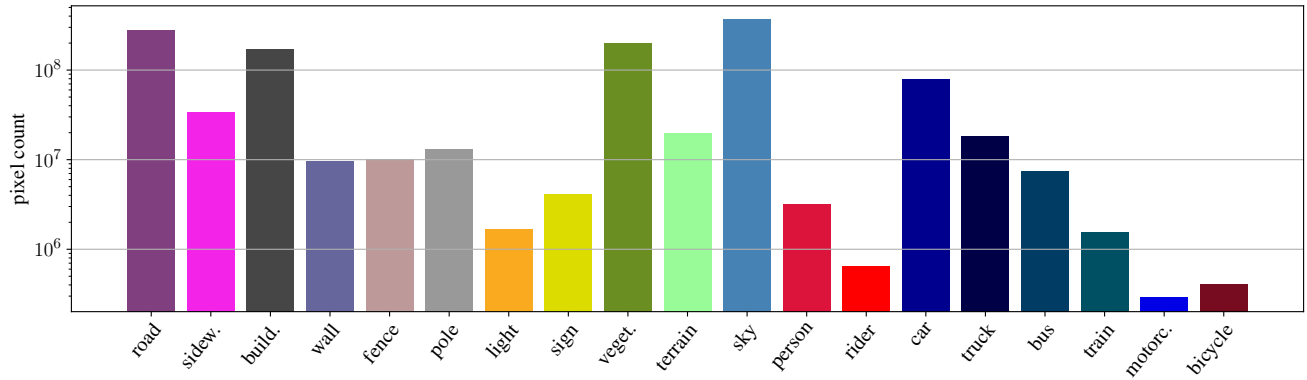}%
    }%
    \caption{Number of annotated pixels per class in the ACG benchmark.}
    \label{fig:pixel_counts}
\end{figure*}

\section{ACG Benchmark}
\label{sec:supp_acg_benchmark}

The purpose of ACG is to provide a generalization benchmark estimating a model's adverse-condition robustness to diverse inputs, whereby the model is trained on another dataset such as ACDC~\cite{sakaridis2021acdc}, Mapillary Vistas~\cite{neuhold2017mapillary}, \etc.
The evaluation benchmark consists of training, validation, and test images from the public datasets Wilddash2~\cite{zendel2022unifying}, BDD100K~\cite{yu2020bdd100k}, Foggy Zurich~\cite{dai2020curriculum}, and Foggy Driving~\cite{sakaridis2018semantic}.
Models trained on these four datasets can therefore not be evaluated on ACG.

\subsection{Construction}

We constructed the ACG benchmark as follows:

\begin{enumerate}
\itemsep0em 

\item For each of Wilddash2, BDD100K, Foggy Driving, and Foggy Zurich, we inspected all images with public semantic segmentation annotations and extracted images depicting fog, night, rain, or snow\textemdash or a combination thereof. For Wilddash2 we only considered images taken in Europe and North America, to confine the geographical domain shift.

\item For every selected image, we checked the quality of the semantic segmentation labels. Images with unreliable ground-truth were eliminated. For BDD100K, this step eliminated a majority of images.

\item For selected images from Foggy Driving or Foggy Zurich, we checked for potential geographical overlap with ACDC, since all three datasets were recorded in the greater area of Zurich. Images with geographical overlap were eliminated.
\end{enumerate}

Through these three steps, we selected 919 images from a pool of 15173.
However, upon closer inspection we observed that there were no rainy images containing the ``train'' class, which would prevent condition-wise evaluation (see Sec.~\ref{subsec:supp_data_split}).
We therefore collected 3 copyright-free images from the web depicting rainy street scenes with trains or trams and finely annotated the pixels of class ``train''.
In total, ACG consists of 922 adverse-condition images with high-quality ground-truth annotations.

The ground-truth annotations follow the labeling convention of Cityscapes~\cite{cordts2016cityscapes}, consisting of 19 classes. For Wilddash2, semantic classes were mapped back to Cityscapes classes according to the mapping given by~\cite{zendel2022unifying}.

\subsection{Data Splits}
\label{subsec:supp_data_split}

We divide the 922 images into 4 subsets, classified by condition, to enable condition-wise evaluation.
Each image depicting a nighttime scene was assigned to ACG-night, regardless of the weather condition.
For daytime images, each image was assigned to either ACG-fog, ACG-rain, or ACG-snow, depending on the dominant weather condition.
The resulting subset sizes are 121 for ACG-fog, 225 for ACG-rain, 276 for ACG-snow, and 300 for ACG-night.

\subsection{Class Distribution}

The numbers of annotated pixels per class are shown in Fig.~\ref{fig:pixel_counts}.
Importantly, each class is also represented within each condition-subset.

\beginappendixc
\section{Additional Ablations}

\PAR{Effect of Entropy and Self-Training Losses.}
We show in Table~\ref{tab:aux_losses} the effect of the individual training losses on ACDC validation performance.
Omitting either the self-training or our CDC loss leads to a large performance drop, while omitting the entropy loss has a more minor effect.

\begin{table}
\caption{Effect of individual training losses on ACDC validation performance.}%
\smallskip%
\centering%
\resizebox{\columnwidth}{!}{%
\ra{1.3}%
\setlength{\tabcolsep}{20pt}%
\begin{tabular}{@{}lcccr@{}}\toprule
& CMA & w/o $\mathcal{L}_\text{ent}$ & w/o $\mathcal{L}_\text{st}$ & w/o $\mathcal{L}_\text{cdc}$ \\
\midrule 
ACDC val mIoU & 67.2 & 66.7 & 57.7 & 60.1 \\
\bottomrule
\end{tabular} }%
\label{tab:aux_losses}
\end{table}

\PAR{Sensitivity to Patch Confidence Threshold.}
Table~\ref{tab:sensitivity} shows the sensitivity of CMA to the confidence threshold value, which is set to 0.2 in Eq.~\eqref{eq:discarding}.

\begin{table}
\caption{Sensitivity of CMA to the confidence threshold (default value of 0.2).}%
\smallskip%
\centering%
\resizebox{\columnwidth}{!}{%
\ra{1.3}%
\setlength{\tabcolsep}{12pt}%
\begin{tabular}{@{}lcccccr@{}}
    \toprule
    confidence threshold & 0 & 0.1 & 0.2 & 0.3 & 0.4 & 0.5 \\
    \midrule
    ACDC val mIoU & 66.8 & 67.1 & 67.2 & 67.0 & 67.0 & 66.7 \\
    \bottomrule
\end{tabular}}%
\label{tab:sensitivity}
\end{table}

\beginappendixd
\section{Condition-Wise ACDC Performances}

In Tables~\ref{tab:acdc_fog}, \ref{tab:acdc_night}, \ref{tab:acdc_rain}, \ref{tab:acdc_snow} we report the test set results for the condition-wise evaluations on ACDC-fog, ACDC-night, ACDC-rain, and ACDC-snow. 
For all the methods, Cityscapes is used as the source dataset and the full ACDC training set as the target dataset.
On ACDC-night, ACDC-rain, and ACDC-snow, CMA outperforms all other methods, while being second best on ACDC-fog.

\begin{table*}[!p]
\caption{Comparison to the state of the art in model adaptation on Cityscapes$\to$ACDC, with reported performance on the ACDC-fog test set.}%
\smallskip%
\centering%
\resizebox{\linewidth}{!}{%
\ra{1.3}%
\begin{tabular}{@{}lccccccccccccccccccccccr@{}}\toprule
\multirow{2}{*}{Method} &&& \multicolumn{21}{c}{ACDC-fog IoU\,$\uparrow$} \\
\cmidrule{4-24} &&& \rotatebox[origin=c]{90}{road} & \rotatebox[origin=c]{90}{sidew.} & \rotatebox[origin=c]{90}{build.} & \rotatebox[origin=c]{90}{wall} & \rotatebox[origin=c]{90}{fence} & \rotatebox[origin=c]{90}{pole} & \rotatebox[origin=c]{90}{light} & \rotatebox[origin=c]{90}{sign} & \rotatebox[origin=c]{90}{veget.} & \rotatebox[origin=c]{90}{terrain} & \rotatebox[origin=c]{90}{sky} & \rotatebox[origin=c]{90}{person} & \rotatebox[origin=c]{90}{rider} & \rotatebox[origin=c]{90}{car} & \rotatebox[origin=c]{90}{truck} & \rotatebox[origin=c]{90}{bus} & \rotatebox[origin=c]{90}{train} & \rotatebox[origin=c]{90}{motorc.} & \rotatebox[origin=c]{90}{bicycle} && \multicolumn{1}{c}{\phantom{00}\rotatebox[origin=c]{90}{\textbf{mean}}} \\ \midrule
Source model & \multirow{5}{*}{\rotatebox[origin=c]{90}{SegFormer~\cite{xie2021segformer}}} && 87.8 & 60.7 & 73.1 & 44.5 & 30.1 & 42.1 & 52.3 & 64.4 & 81.4 & 68.8 & 93.4 & 51.1 & 53.2 & 78.4 & 66.0 & 39.7 & 75.1 & 43.2 & \textbf{47.4} && 60.7 \\
TENT~\cite{wang2021tent} &&& 83.0 & 61.1 & 68.2 & 44.1 & 30.4 & 44.1 & 52.1 & 64.4 & 81.1 & 69.3 & 89.9 & 50.9 & 54.7 & 78.6 & 67.1 & 39.5 & 75.4 & 45.9 & 47.1 && 60.4 \\
HCL~\cite{huang2021model} &&& 88.5 & 63.2 & 79.8 & 45.3 & 30.6 & 44.7 & 53.7 & 65.9 & 81.8 & 69.6 & 95.5 & 52.5 & 55.0 & 79.4 & 68.0 & 40.7 & 74.0 & 40.7 & 46.9 && 61.9 \\
URMA~\cite{fleuret2021uncertainty} &&& 89.3 & 61.8 & 87.9 & 51.4 & \textbf{36.3} & \textbf{52.3} & 58.1 & \textbf{67.9} & 85.7 & \textbf{71.8} & 97.2 & 54.5 & \textbf{62.5} & \textbf{82.3} & \textbf{70.6} & \textbf{62.0} & 82.0 & \textbf{52.9} & 36.2 && \textbf{66.5} \\
CMA &&& \textbf{93.5} & \textbf{75.3} & \textbf{88.6} & \textbf{53.4} & 33.0 & 52.2 & \textbf{58.2} & 67.0 & \textbf{86.9} & 71.5 & \textbf{97.8} & \textbf{55.6} & 42.0 & 80.4 & 70.0 & 54.8 & \textbf{83.3} & 43.0 & 37.4 && 65.5 \\
\bottomrule
\end{tabular}}%
\label{tab:acdc_fog}
\end{table*}

\begin{table*}[!p]
\caption{Comparison to the state of the art in model adaptation on Cityscapes$\to$ACDC, with reported performance on the ACDC-night test set.}%
\smallskip%
\centering%
\resizebox{\linewidth}{!}{%
\ra{1.3}%
\begin{tabular}{@{}lccccccccccccccccccccccr@{}}\toprule
\multirow{2}{*}{Method} &&& \multicolumn{21}{c}{ACDC-night IoU\,$\uparrow$} \\
\cmidrule{4-24} &&& \rotatebox[origin=c]{90}{road} & \rotatebox[origin=c]{90}{sidew.} & \rotatebox[origin=c]{90}{build.} & \rotatebox[origin=c]{90}{wall} & \rotatebox[origin=c]{90}{fence} & \rotatebox[origin=c]{90}{pole} & \rotatebox[origin=c]{90}{light} & \rotatebox[origin=c]{90}{sign} & \rotatebox[origin=c]{90}{veget.} & \rotatebox[origin=c]{90}{terrain} & \rotatebox[origin=c]{90}{sky} & \rotatebox[origin=c]{90}{person} & \rotatebox[origin=c]{90}{rider} & \rotatebox[origin=c]{90}{car} & \rotatebox[origin=c]{90}{truck} & \rotatebox[origin=c]{90}{bus} & \rotatebox[origin=c]{90}{train} & \rotatebox[origin=c]{90}{motorc.} & \rotatebox[origin=c]{90}{bicycle} && \multicolumn{1}{c}{\phantom{00}\rotatebox[origin=c]{90}{\textbf{mean}}} \\ \midrule
Source model & \multirow{5}{*}{\rotatebox[origin=c]{90}{SegFormer~\cite{xie2021segformer}}} && 87.9 & 52.7 & 64.1 & 34.0 & 20.2 & 37.2 & 34.5 & 40.2 & 51.8 & 32.4 & \phantom{0}6.6 & 54.5 & 31.4 & 72.8 & 49.6 & 65.2 & 54.1 & 34.0 & 41.4 && 45.5 \\
TENT~\cite{wang2021tent} &&& 85.9 & 53.3 & 64.3 & 34.4 & 20.2 & 37.8 & 35.2 & 40.3 & 52.3 & 33.9 & \phantom{0}2.9 & 53.8 & 31.9 & 72.5 & 46.2 & 63.8 & 53.8 & 34.0 & 40.9 && 45.1 \\
HCL~\cite{huang2021model} &&& 88.2 & 54.3 & 64.4 & 35.3 & 20.7 & 39.1 & 36.8 & 40.4 & 52.0 & 32.1 & \phantom{0}2.8 & 55.2 & 33.7 & 73.5 & 49.2 & 66.5 & 58.1 & 35.4 & 41.7 && 46.3 \\
URMA~\cite{fleuret2021uncertainty} &&& 90.6 & 60.1 & 71.9 & 42.6 & 26.7 & 47.5 & 47.5 & 47.4 & 46.7 & 42.9 & \phantom{0}0.4 & 54.4 & 34.6 & 76.8 & 42.1 & 65.6 & 71.0 & \textbf{38.0} & 37.2 && 49.7 \\
CMA &&& \textbf{95.2} & \textbf{77.5} & \textbf{84.3} & \textbf{43.9} & \textbf{30.9} & \textbf{49.4} & \textbf{52.0} & \textbf{49.6} & \textbf{74.2} & \textbf{51.2} & \textbf{78.4} & \textbf{61.4} & \textbf{41.2} & \textbf{79.2} & \textbf{63.6} & \textbf{75.1} & \textbf{75.8} & 34.6 & \textbf{47.3} && \textbf{61.3} \\
\bottomrule
\end{tabular}}%
\label{tab:acdc_night}
\end{table*}

\begin{table*}[!p]
\caption{Comparison to the state of the art in model adaptation on Cityscapes$\to$ACDC, with reported performance on the ACDC-rain test set.}%
\smallskip%
\centering%
\resizebox{\linewidth}{!}{%
\ra{1.3}%
\begin{tabular}{@{}lccccccccccccccccccccccr@{}}\toprule
\multirow{2}{*}{Method} &&& \multicolumn{21}{c}{ACDC-rain IoU\,$\uparrow$} \\
\cmidrule{4-24} &&& \rotatebox[origin=c]{90}{road} & \rotatebox[origin=c]{90}{sidew.} & \rotatebox[origin=c]{90}{build.} & \rotatebox[origin=c]{90}{wall} & \rotatebox[origin=c]{90}{fence} & \rotatebox[origin=c]{90}{pole} & \rotatebox[origin=c]{90}{light} & \rotatebox[origin=c]{90}{sign} & \rotatebox[origin=c]{90}{veget.} & \rotatebox[origin=c]{90}{terrain} & \rotatebox[origin=c]{90}{sky} & \rotatebox[origin=c]{90}{person} & \rotatebox[origin=c]{90}{rider} & \rotatebox[origin=c]{90}{car} & \rotatebox[origin=c]{90}{truck} & \rotatebox[origin=c]{90}{bus} & \rotatebox[origin=c]{90}{train} & \rotatebox[origin=c]{90}{motorc.} & \rotatebox[origin=c]{90}{bicycle} && \multicolumn{1}{c}{\phantom{00}\rotatebox[origin=c]{90}{\textbf{mean}}} \\ \midrule
Source model & \multirow{5}{*}{\rotatebox[origin=c]{90}{SegFormer~\cite{xie2021segformer}}} && 83.1 & 46.7 & 89.5 & 40.5 & 47.2 & 54.0 & 67.0 & 66.9 & 92.6 & 40.2 & 97.6 & 63.5 & 24.6 & 87.8 & 65.1 & 72.7 & 81.0 & 42.8 & 58.0 && 64.3 \\
TENT~\cite{wang2021tent} &&& 83.1 & 47.2 & 89.2 & 40.9 & 47.6 & 54.5 & 66.9 & 67.3 & 92.7 & 41.4 & 97.1 & 63.7 & 25.4 & 87.9 & 65.3 & 74.8 & 82.2 & 43.1 & 57.4 && 64.6 \\
HCL~\cite{huang2021model} &&& 84.2 & 50.5 & 90.1 & 42.7 & 48.9 & 57.0 & 68.5 & 69.0 & 93.0 & 40.9 & 97.8 & 65.4 & 26.1 & 88.7 & 68.1 & 74.4 & 80.4 & 43.8 & 58.0 && 65.6 \\
URMA~\cite{fleuret2021uncertainty} &&& 87.2 & 61.0 & 92.4 & 52.0 & 51.9 & 57.2 & \textbf{72.0} & \textbf{73.1} & \textbf{93.8} & \textbf{46.1} & \textbf{98.1} & \textbf{68.8} & 31.8 & \textbf{90.6} & \textbf{73.2} & 85.9 & \textbf{86.9} & \textbf{51.7} & 51.9 && 69.8 \\
CMA &&& \textbf{93.3} & \textbf{76.3} & \textbf{92.8} & \textbf{58.1} & \textbf{58.2} & \textbf{61.2} & 70.4 & 71.8 & \textbf{93.8} & 45.0 & 97.9 & 67.4 & \textbf{36.8} & 89.7 & 72.2 & \textbf{88.5} & 86.4 & 50.5 & \textbf{66.7} && \textbf{72.5} \\
\bottomrule
\end{tabular}}%
\label{tab:acdc_rain}
\end{table*}

\begin{table*}[!p]
\caption{Comparison to the state of the art in model adaptation on Cityscapes$\to$ACDC, with reported performance on the ACDC-snow test set.}%
\smallskip%
\centering%
\resizebox{\linewidth}{!}{%
\ra{1.3}%
\begin{tabular}{@{}lccccccccccccccccccccccr@{}}\toprule
\multirow{2}{*}{Method} &&& \multicolumn{21}{c}{ACDC-snow IoU\,$\uparrow$} \\
\cmidrule{4-24} &&& \rotatebox[origin=c]{90}{road} & \rotatebox[origin=c]{90}{sidew.} & \rotatebox[origin=c]{90}{build.} & \rotatebox[origin=c]{90}{wall} & \rotatebox[origin=c]{90}{fence} & \rotatebox[origin=c]{90}{pole} & \rotatebox[origin=c]{90}{light} & \rotatebox[origin=c]{90}{sign} & \rotatebox[origin=c]{90}{veget.} & \rotatebox[origin=c]{90}{terrain} & \rotatebox[origin=c]{90}{sky} & \rotatebox[origin=c]{90}{person} & \rotatebox[origin=c]{90}{rider} & \rotatebox[origin=c]{90}{car} & \rotatebox[origin=c]{90}{truck} & \rotatebox[origin=c]{90}{bus} & \rotatebox[origin=c]{90}{train} & \rotatebox[origin=c]{90}{motorc.} & \rotatebox[origin=c]{90}{bicycle} && \multicolumn{1}{c}{\phantom{00}\rotatebox[origin=c]{90}{\textbf{mean}}} \\ \midrule
Source model & \multirow{5}{*}{\rotatebox[origin=c]{90}{SegFormer~\cite{xie2021segformer}}} && 82.0 & 44.9 & 80.5 & 30.4 & 45.4 & 46.8 & 65.6 & 63.1 & 86.8 & \phantom{0}5.2 & 93.6 & 67.8 & 40.8 & 87.1 & 56.4 & 76.7 & 83.1 & 32.8 & 60.3 && 60.5 \\
TENT~\cite{wang2021tent} &&& 81.8 & 45.6 & 79.1 & 31.3 & 45.4 & 48.0 & 65.5 & 63.3 & 86.9 & \phantom{0}4.6 & 91.8 & 67.4 & 43.1 & 87.0 & 53.3 & 76.6 & 83.2 & 33.6 & 61.9 && 60.5 \\
HCL~\cite{huang2021model} &&& 82.9 & 47.4 & 83.2 & 35.4 & 46.8 & 50.1 & 67.8 & 64.9 & 87.7 & \phantom{0}5.3 & 95.6 & 69.8 & 43.9 & 87.6 & 60.1 & 76.9 & 83.2 & 35.3 & 63.4 && 62.5 \\
URMA~\cite{fleuret2021uncertainty} &&& 88.0 & 58.9 & 87.2 & \textbf{52.0} & 51.7 & \textbf{57.8} & \textbf{75.6} & 70.3 & 88.8 & \phantom{0}5.8 & \textbf{97.1} & 75.0 & \textbf{63.6} & 89.0 & \textbf{69.6} & 79.0 & \textbf{89.8} & \textbf{50.1} & 65.4 && 69.2 \\
CMA &&& \textbf{92.4} & \textbf{70.5} & \textbf{88.3} & 50.4 & \textbf{55.6} & 56.3 & 74.8 & \textbf{71.1} & \textbf{90.8} & \textbf{29.4} & 96.9 & \textbf{77.4} & 63.5 & \textbf{90.1} & 63.5 & \textbf{79.6} & 89.0 & 45.6 & \textbf{73.9} && \textbf{71.5} \\
\bottomrule
\end{tabular}}%
\label{tab:acdc_snow}
\end{table*}

\beginappendixe
\section{Source Model Predictions}
\label{sec:supp_source_preds}

\begin{figure*}
    \adjustbox{max width=\textwidth}{%
        \input{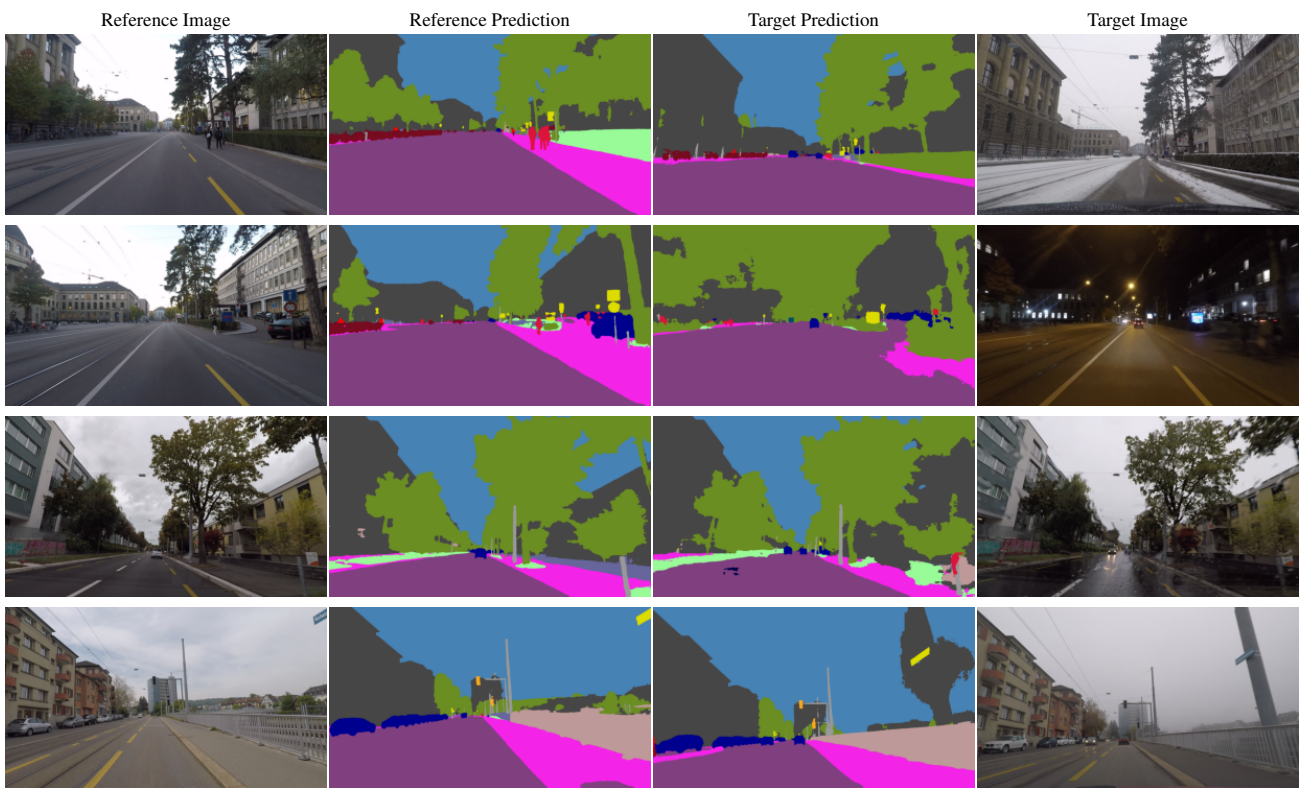}%
    }%
    \caption{Comparison of SegFormer predictions on pairs of reference and target images.}
    \label{fig:quali_segformer}
\end{figure*}

Fig.~\ref{fig:quali_segformer} shows SegFormer source model predictions on corresponding reference (normal condition, left) and target (adverse condition, right) images of the ACDC dataset.
Overall, the Cityscapes-trained source model produces more accurate predictions on the reference images.

\beginappendixf
\section{Qualitative Results}
\label{sec:supp_qualitative_res}

\begin{figure*}
    \adjustbox{max width=\textwidth}{%
        \input{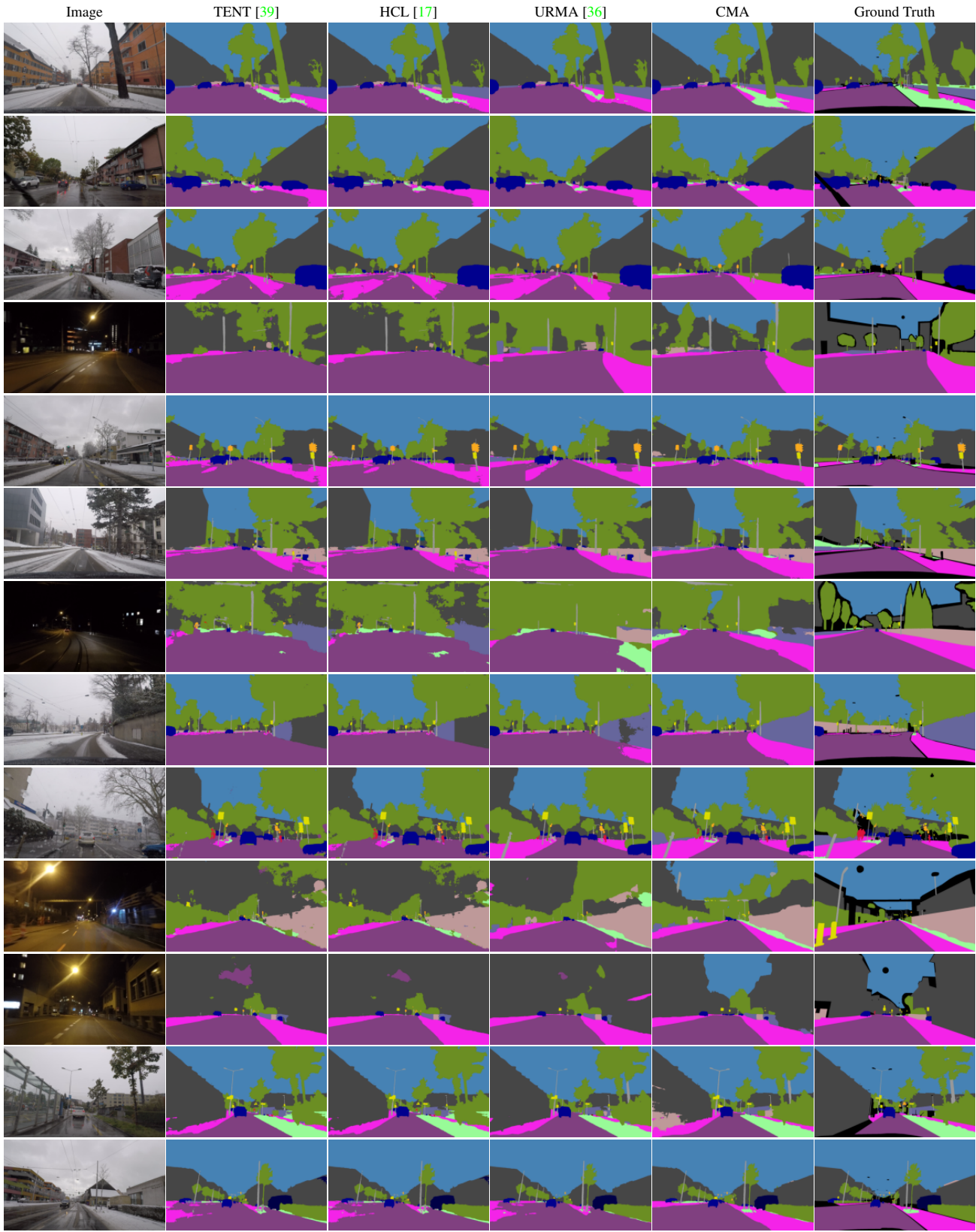}%
    }%
    \caption{Qualitative segmentation results of SegFormer-based adaptation methods on ACDC validation images.}
    \label{fig:sup_qualitative_seg}
\end{figure*}

We provide more qualitative segmentation results on randomly selected ACDC validation images in Fig.~\ref{fig:sup_qualitative_seg}. 

\end{appendices}

\end{document}